%% file: neurips_2025.tex
\documentclass[table, html]{article}
\PassOptionsToPackage{numbers}{natbib}

    \usepackage[preprint]{neurips_2025}

\usepackage[utf8]{inputenc} 
\usepackage[T1]{fontenc}    
\usepackage{hyperref}       
\usepackage{url}            
\usepackage{booktabs}       
\usepackage{amsfonts}       
\usepackage{nicefrac}       
\usepackage{microtype}      
\usepackage{graphicx}
\usepackage{subcaption}
\usepackage{float}
\usepackage{multirow}
\usepackage{multicol}
\usepackage{tcolorbox}
\usepackage{verbatim}
\usepackage{fancyvrb}
\usepackage{wrapfig}
\usepackage{xcolor}
\usepackage{amsmath,amsfonts,bm}

\newcommand{\bench}{\textsc{TMGBench}}

\definecolor{goldenyellow}{RGB}{255, 223, 0}
\definecolor{grassgreen}{RGB}{124, 252, 0}

\title{TMGBench: A Systematic Game Benchmark for Evaluating Strategic Reasoning Abilities of LLMs}

\author{
  Haochuan Wang \\
  \texttt{wanghc03@stu.hit.edu.cn}
  \And
  Xiachong Feng \\
  \texttt{xiachongfeng1996@gmail.com}
  \AND
  Lei Li \\
  \texttt{nlp.lilei@gmail.com}
  \And
  Yu Guo \\
  \texttt{yuguo\_hit@163.com}
  \And
  Zhanyue Qin \\
  \texttt{zhanyueqin@stu.hit.edu.cn}
  \AND
  Dianbo Sui \\
  \texttt{suidianbo@hit.edu.cn}
  \And
  Lingpeng Kong \\
  \texttt{lpk@cs.hku.hk}
}

\begin{document}

\maketitle

\begin{abstract}
The rapid advancement of large language models (LLMs) has accelerated their application in reasoning, with strategic reasoning drawing increasing attention.
To evaluate the strategic reasoning capabilities of LLMs, game theory, with its concise structure, has become the preferred approach for many researchers.
However, current research typically focuses on a limited selection of games, resulting in low coverage of game types. 
Additionally, classic game scenarios carry risks of data leakage, and the benchmarks used often lack extensibility, rendering them inadequate for evaluating state-of-the-art models.
To address these challenges, we propose {\bench}, a benchmark characterized by comprehensive game type coverage, novel and diverse scenarios, and flexible game organization. 
Specifically, we incorporate all 144 game types summarized by the Robinson-Goforth topology of 2×2 games, which are constructed as classic games in our benchmark. 
Furthermore, we employ synthetic data generation techniques to create diverse, higher-quality game scenarios through topic guidance and human inspection for each classic game, which we refer to as story-based games.
Lastly, to provide a sustainable evaluation framework adaptable to increasingly powerful LLMs, we treat the aforementioned games as atomic units and organize them into more complex forms through sequential, parallel, and nested structures.
We conducted a comprehensive evaluation of mainstream LLMs, covering tests on rational reasoning, reasoning robustness, Theory-of-Mind capabilities, and reasoning in complex game forms. 
The results revealed LLMs still have flaws in the accuracy and consistency of strategic reasoning processes, and their levels of mastery over Theory-of-Mind also vary.
Additionally, SOTA models like o3-mini, Qwen3 and deepseek-reasoner, were also evaluated across the sequential, parallel, and nested game structures while the results highlighted the challenges posed by {\bench}.
\end{abstract}

\input{sections/introduction}
\input{sections/benchmark}
\input{sections/analysis}
\input{sections/related_work}
\input{sections/discussion}

\bibliography{neurips_2025}
\bibliographystyle{neurips_2025}

\newpage
\appendix
\input{sections/appendix}

\end{document}

%% file: sections/introduction.tex
\section{Introduction}
The rapid advancement of large language models (LLMs) has reshaped the paradigm of artificial intelligence, achieving breakthroughs across various domains~\citep{jaech2024openai,qwen2.5,guo2025deepseek,team2025gemma,guo2025seed15vltechnicalreport}.
These achievements are largely attributed to LLMs' ability to assimilate vast amounts of knowledge during training, emerging with the capacity to organize information at a coarse level and link knowledge at a fine-grained level through their internal representations~\citep{wang2024knowledge}. 
These core capabilities have driven the success of LLMs in numerous reasoning tasks, including mathematical reasoning~\citep{shao2024deepseekmath}, commonsense reasoning~\citep{toroghi-etal-2024-verifiable}, logical reasoning~\citep{lei2023boosting}, and strategic reasoning~\citep{lore2023strategic, duan2024gtbench}. 
Among these, strategic reasoning has attracted considerable attention due to its multi-agent nature and close association with social intelligence~\citep{gandhi2023strategic,feng2024survey}.

Strategic reasoning refers to the cognitive process of anticipating, planning, and responding to others' actions to achieve specific objectives within competitive or cooperative contexts~\citep{zhang2024llm}. 
Consequently, game scenarios—naturally involving both cooperation and competition—have intuitively become a fertile ground for studying LLMs' strategic reasoning abilities~\citep{brookins2023playing}. 
In particular, researchers have engaged LLMs in game-playing, analyzing their decision-making behaviors and evaluating their strategic intelligence in such scenarios~\citep{duan2024gtbench}. 
The Prisoner's Dilemma, as one of the most classic game theory scenarios, has been extensively studied in this context~\citep{herr2024large}. 
Additionally, other traditional games such as the Battle of the Sexes~\citep{kreps1990game}, the Stag Hunt~\citep{carlsson199312}, and the Dictator Game~\citep{forsythe1994fairness} have also drawn significant attention. 
These studies provide initial insights into the strategic reasoning capabilities of LLMs~\citep{horton2023large, phelps2023investigating, akata2023playing, li2023beyond,hua2024game,shapira2024glee}.

However, current research has three major limitations, hindering a comprehensive, robust, and sustainable evaluation of LLMs' strategic reasoning capabilities:
(1) \textit{Limited coverage of game types}: Most studies focus on a handful of classic games without considering the full diversity of game structures.
(2) \textit{Potential risk of game scenario leakage}: Classic game scenarios are likely to be present in the training corpus, raising concerns over data leakage.
(3) \textit{Poor extensibility of game forms}: Existing studies primarily focus on a narrow range of game forms, which may no longer suffice to challenge high-performing LLMs such as o1-mini from OpenAI.

To address the above issues, we introduce {\bench}, a benchmark that encompasses a comprehensive range of game types, features synthesized game scenarios, and supports scalable and reorganizable game forms. 
Specifically, \textit{to address the first issue}, we include all 144 game types defined by the Robinson-Goforth topology of 2x2 games~\citep{robinson2005topology}. 
This topology encompasses a variety of game structures based on different numerical payoff matrices, including but not limited to classic games like the Prisoner’s Dilemma(\S\ref{sec:gt}). 
\textit{To address the second issue}, we employ synthetic data generation techniques to create five different story-based games for each classic game.
In essence, a story-based game is a contextual framing counterpart of its corresponding classic game, sharing the same structure but differing in context~\citep{lore2023strategic}. 
To ensure high-quality data synthesis, we introduce two additional steps: topic control and human inspection. 
We first define a set of topics commonly associated with cooperation and competition, such as business and law, to guide the data generation process. 
Then, to ensure that the synthesized games meet the required game structures and are easily understandable, we conduct rigorous human inspection (\S\ref{sec:cf}).
\textit{To address the third issue}, we propose three forms for expanding and organizing games: sequential, parallel, and nested. 
Using the above constructed games as atomic units, we reorganize them into these complex forms to assess the strategic reasoning of LLMs. 
The sequential and parallel forms evaluate the model's capacity for sequential and parallel decision-making, respectively, while the nested form explores the LLMs' multi-layered strategic reasoning abilities (\S\ref{sec:cxf}).

Based on {\bench}, we conduct comprehensive analyses and evaluations of current mainstream LLMs (\S\ref{sec:exp}), including assessments of rational reasoning, reasoning robustness, Theory-of-Mind (ToM) capabilities, and reasoning in complex game forms, leading to the following key findings:

(1) Advanced LLMs like o3-mini, Qwen3, deepseek-reasoner,  demonstrate strong strategic reasoning, with over 90\% accuracy. However, other models struggle to generalize across contexts and scenarios in different ways.

(2) Though GPT series models often perform well, their reasoning inconsistency on certain task sub-types is marked by an `asymmetric pattern' as reported, which is the main cause of the statistical biases.

(3) Several LLMs demonstrate stable first-order ToM abilities, with some effectively utilizing second-order ToM for comparable tasks. 
In contrast, models such as Llama-3.1-70B appear restricted to first-order reasoning.

(4) Complex-form games that are derived from atomic units in {\bench} present considerable challenges for LLMs, including those with strong reasoning abilities like o3-mini, which often struggle as the number of games increases.

%% file: sections/benchmark.tex
\section{{\bench}}

\begin{figure}[t]
    \centering
    \begin{minipage}[b]{0.47\textwidth}
        \centering
        \includegraphics[width=\textwidth]{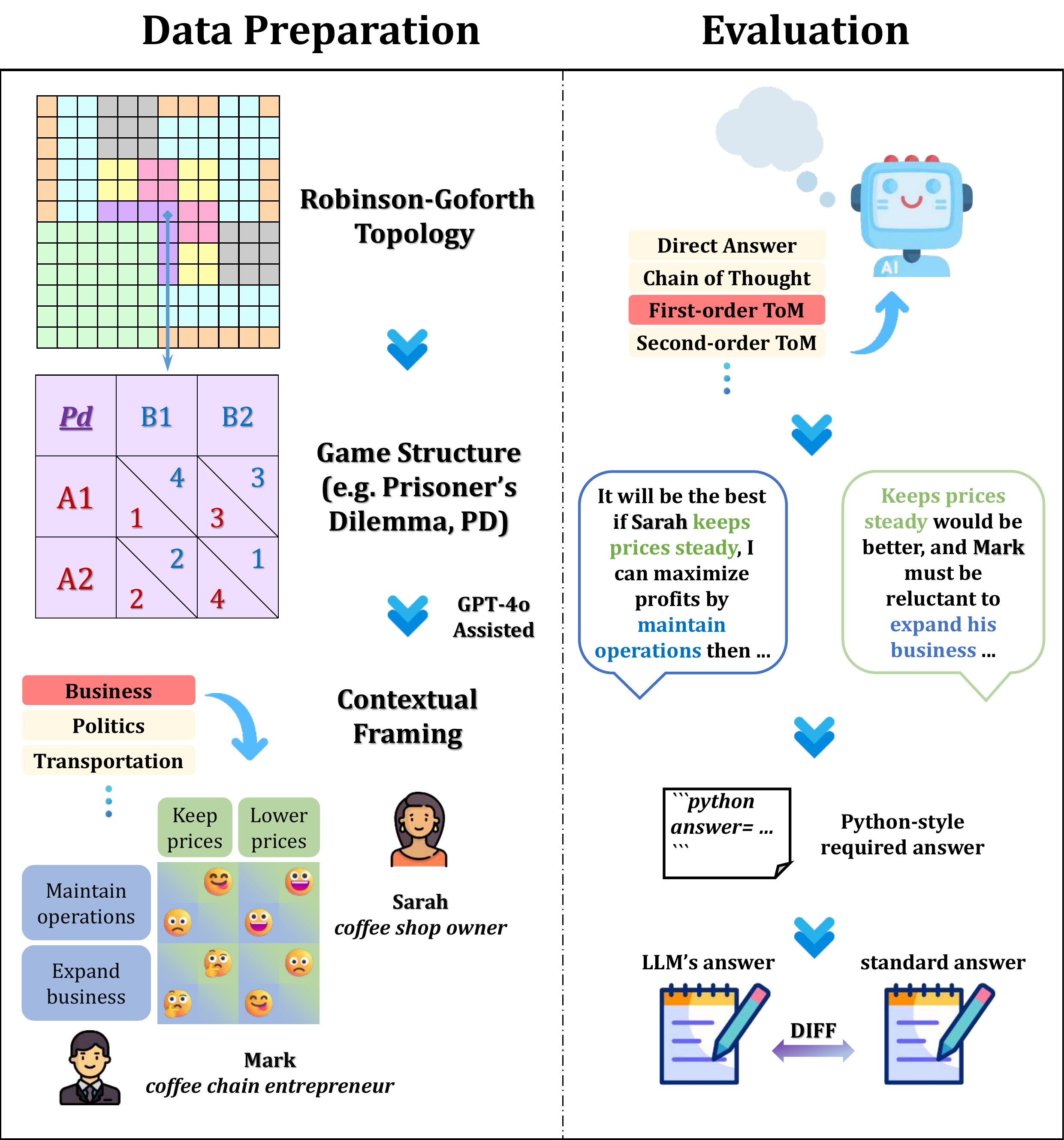}
        \caption{An concept map of {\bench}. The data preparation of the benchmark includes 3 ingredients: Robinson-Goforth topology, game structure and contextual framing.
        The evaluation of the benchmark embraces several prompting methods (including ToM promptings) to elicit strategic reasoning process of LLMs.}
        \label{fig:instance}
    \end{minipage}
    \hspace{0.5cm}
    \begin{minipage}[b]{0.48\textwidth}
        \centering
        \includegraphics[width=\textwidth]{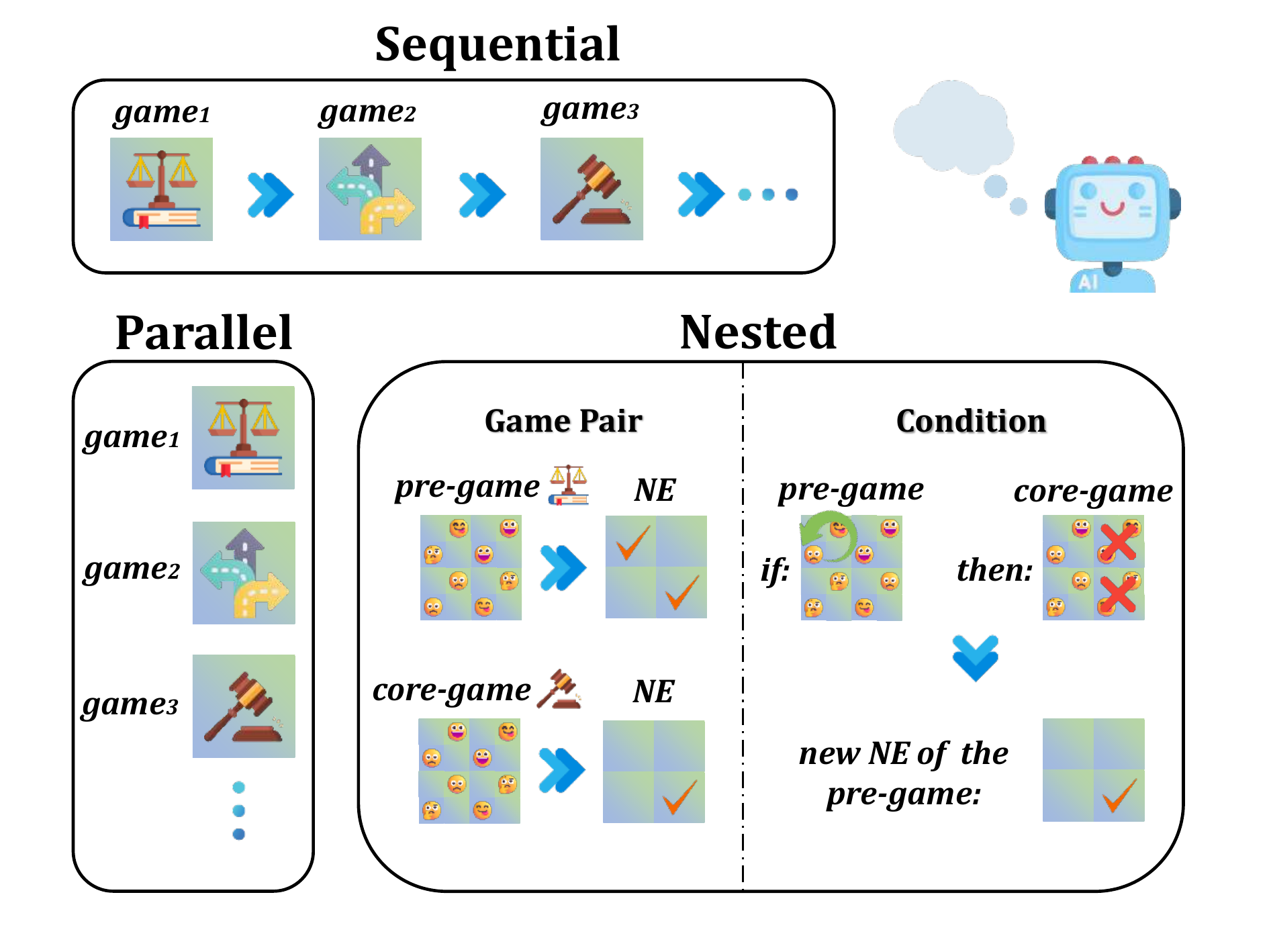}
        \caption{We design several complex forms of strategic reasoning tasks using {\bench}.
        which include: (1) sequential form, where LLMs are required to respond to multiple game tasks in a row, with history of previous tasks; (2) parallel form, where LLMs are required to response multiple game tasks simultaneously; (3) nested form, where LLMs are required to response a set of interlinked game tasks (in our settings, we relate to them as \texttt{pre-game} and \texttt{core-game}). Games in the complex forms can be selected with different game structures and various contexts.}
        \label{fig:benchmark:complex_scenario}
    \end{minipage}
\end{figure}

\subsection{Benchmark Overview}\label{sec:bo}
{\bench} is a benchmark designed to evaluate the strategic reasoning capabilities of LLMs in game-theoretic scenarios, illustrated by Figure~\ref{fig:instance}.
It comprehensively covers 144 types of games (see \S\ref{sec:gt}), with each type containing multiple instances (in each instance, there are two players and each player can choose between two strategies, resulting in four possible situations), which can be categorized into classic and story-based settings. 
Notably, the story-based instances are produced using synthetic data generation techniques and are grounded in real-life themes, effectively mitigating the issue of data leakage (see \S\ref{sec:cf}). Furthermore, each game in {\bench} can be treated as an atomic unit, and multiple atomic games can be structured in a more complex task with parallel, sequential, or nested form (see \S\ref{sec:cxf}). These complex scenarios effectively facilitate the evaluation of advanced LLMs' abilities in parallel, sequential, and multi-layered decision-making. To precisely evaluate the reasoning abilities of LLMs, we use their performance in inferring the optimal strategy combination, i.e., the Nash equilibrium, as the evaluation criterion. Additionally, the designed evaluation metrics provide a fine-grained assessment of the robustness and self-consistency of LLMs' strategic reasoning abilities (see \S\ref{sec:eval}).

\subsection{Game Topology}\label{sec:gt}

Although previous research has explored LLMs' reasoning abilities within the context of game theory, existing studies have primarily focused on a few well-known games, such as the Prisoner's Dilemma, Battle of the Sexes, and Stag Hunt~\citep{brookins2023playing, phelps2023investigating, guo2023gpt}. 
However, these studies cover a limited game types, resulting in incomplete evaluations. 
Thereby, a broader variety of games is urgently needed to conduct a \textit{systematic} assessment of LLMs.

To address this, we incorporate 144 game types (we later refer to a type as an equivalence class) based on the Robinson-Goforth topology of 2×2 games~\citep{robinson2005topology}. 
Classic games like the Prisoner's Dilemma belong to one of the equivalence classes within this topology.
Specifically, the topology of 2×2 games elegantly illustrates the relationships among strictly ordinal 2×2 games, each with a unique payoff structure, leading to different dominant strategies, Nash equilibria, and reasoning approaches (more details in Appendix~\ref{appsec:game:definition}).
We categorize all the 144 games with numerical payoffs from the original topology into the \textit{classic setting} tasks.
Due to space constraints, we provide an introduction to the Robinson-Goforth topology in Appendix~\ref{appsec:game:topology}.

\subsection{Contextual Framing}\label{sec:cf}
Relying on the Robinson-Goforth topology, we can systematically construct all types of classic setting tasks.
However, this alone is insufficient, as games often take place in diverse real-life contexts, involving different topics, types of participants and their preferences. 
Such contextual framing of games introduces new challenges for LLMs~\citep{lore2023strategic}.

To further explore LLMs' strategic reasoning capabilities in real-world scenarios, we use classic games as seed data and employ synthetic data generation techniques, leveraging GPT-4o to construct story-based games.
Specifically, in story-based games, we replace the pure game information of classic games with real-life scenarios, covering topics such as business, law and transportation. Additionally, the two players are substituted with characters representing broader semantics (e.g., people, animals, organizations, and even nations), and the payoff values are transformed from pure numbers into specific states or rewards relevant to the characters. 
For each classic game, we generate 5 corresponding story-based games.

To ensure high-quality data generation, we undertake the following steps:
First, we use GPT-4o to synthesize the contextual data.
Second, we design precise prompts to ensure the generated data adhere to the given game structures.
Third, we select topics from real-life scenarios where strategic interactions are common, guiding the data generation process.
Finally, we conduct rigorous human reviews to ensure the data's quality and diversity. Details on the data generation process, prompts, human review procedures, and topic distribution of the data can be found in Appendix~\ref{appsec:dataset}.

\begin{figure}[t]
    \centering
    \begin{minipage}[b]{0.475\textwidth}
        \centering
        \includegraphics[width=\textwidth]{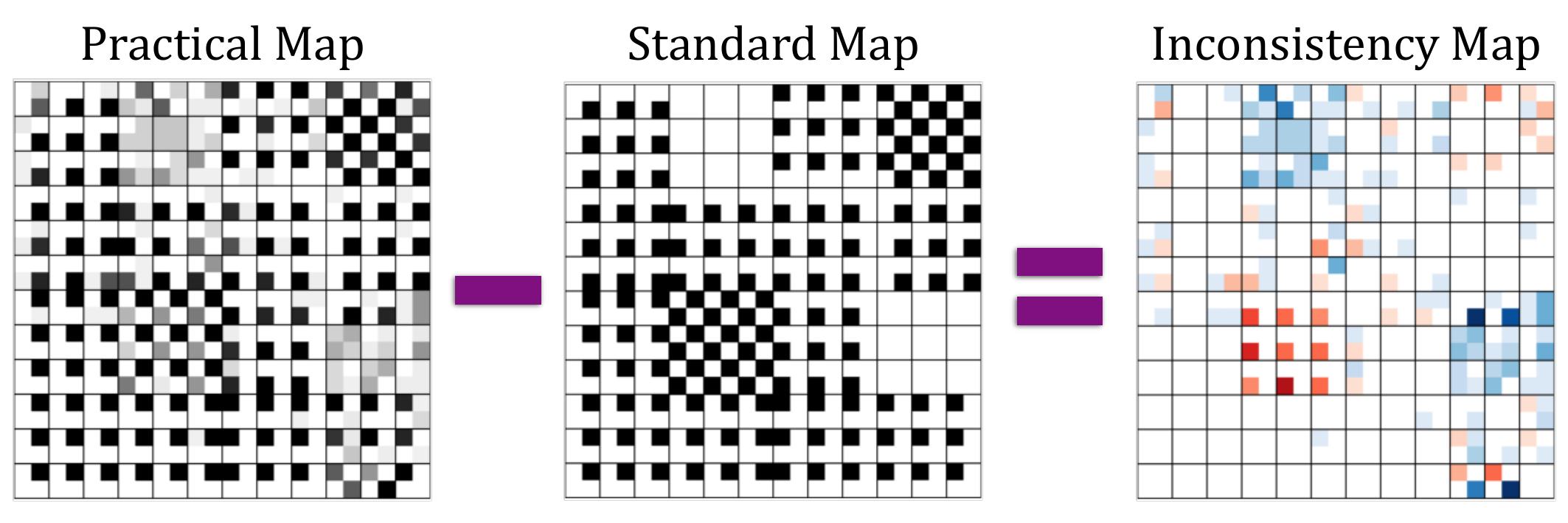}
        \caption{Demonstration of the inconsistency heat map. Each of the grids is divided into 4 quarter-grids, indicating the 4 situations. By subtracting the standard map from the practical map element-wise, we get the inconsistency map, where blue colours indicate positive difference and red colours indicate negative difference. The deeper the colour means the larger the difference between the LLM's response and the standard answer.}
        \label{fig:benchmark:consistency_heat_map}
    \end{minipage}
    \hspace{0.5cm}
    \begin{minipage}[b]{0.475\textwidth}
        \centering
        \includegraphics[width=0.3\textwidth]{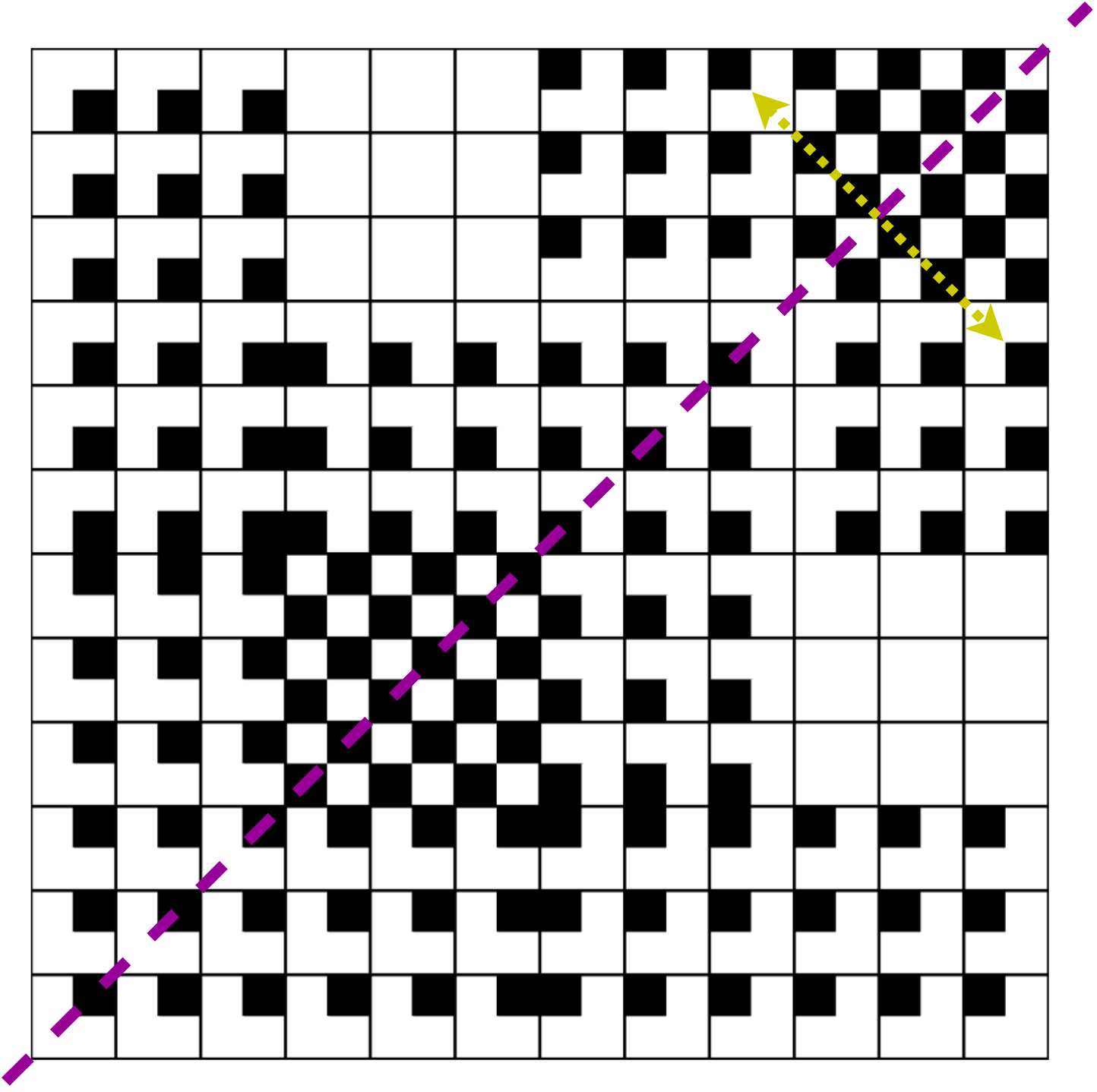}
        \includegraphics[width=0.3\textwidth]{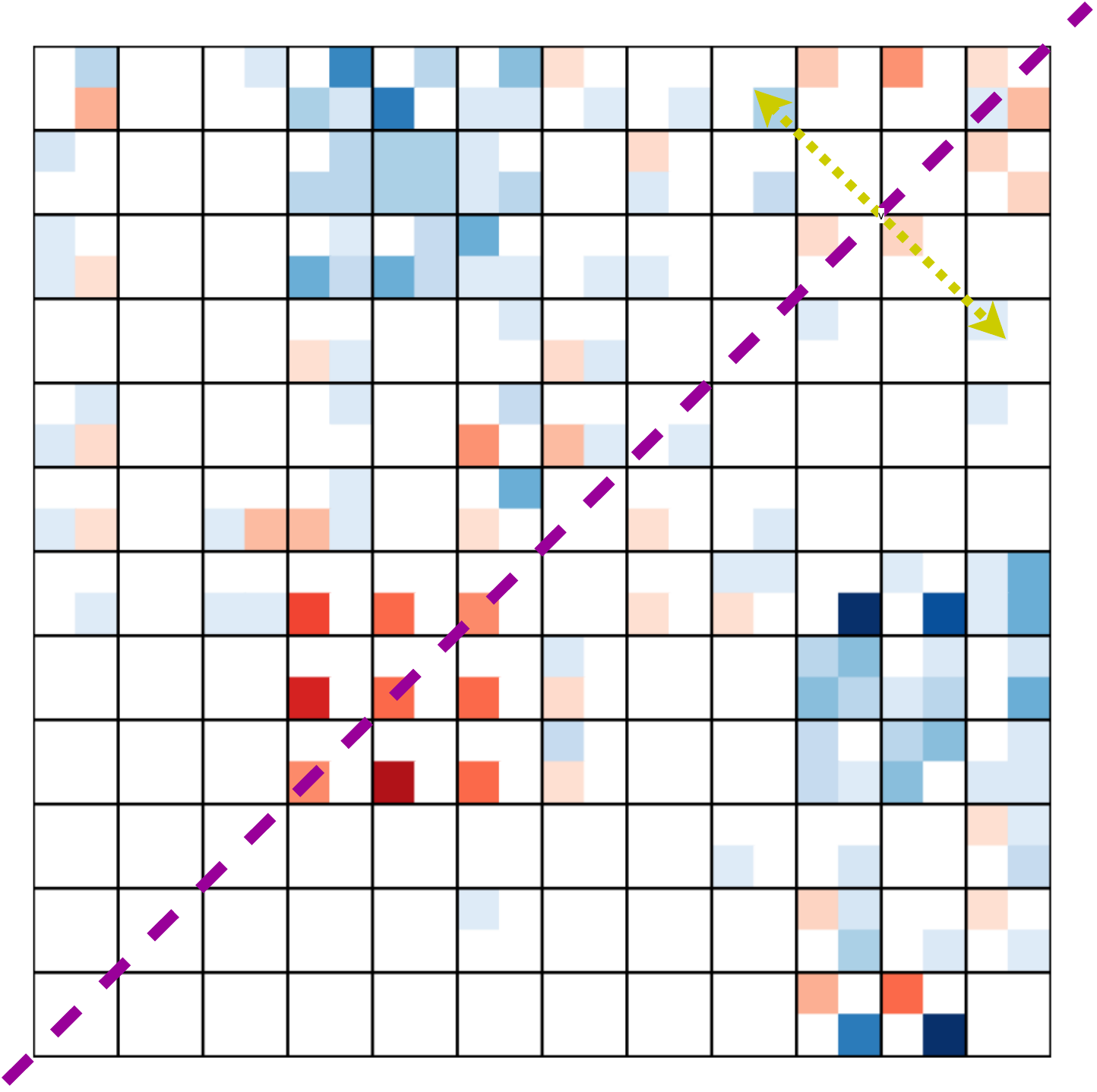}
        \caption{Axisymmetry in heat maps can be illustrated by the left sub-figure, where the standard heat map exhibits perfect axisymmetry across the counter-diagonal. In contrast, LLMs' responses tend to demonstrate quasi-axisymmetry, as shown by the right sub-figure. Certain pairs of positions fail to align precisely when reflected across the axis and may exhibit discrepancies, deviating from the ideal symmetric pattern.}
        \label{fig:benchmark:bias_score}
    \end{minipage}
\end{figure}

\subsection{Complex Forms}\label{sec:cxf}
The 2×2 games in the topology represent a highly condensed game structure. 
However, in real life, we often encounter more complex game forms, such as making continuous decisions, making multiple decisions simultaneously, or considering the impacts of one decision on another.

To evaluate LLMs' strategic reasoning abilities with more constraints, we treat the aforementioned individual games as atomic games and expand them in three forms: sequential, parallel, and nested. 
The organization of these forms is illustrated in Figure~\ref{fig:benchmark:complex_scenario}. Specifically, in the \textit{sequential form}, we randomly sample multiple games from the story-based games, requiring the LLM to make decisions sequentially. 
Only if the LLM provides correct answers for all games is it considered to have made correct decisions. 
In the \textit{parallel form}, the LLM is given multiple randomly sampled games and must make decisions simultaneously. Similarly, the LLM is deemed to have made correct decisions only if it solves all games correctly. In the \textit{nested form}, we randomly sample two games, designated as the \texttt{pre-game} and the \texttt{core-game}, where the \texttt{core-game} holds greater importance. The decisions made by the LLM in the \texttt{pre-game} affect the strategy space in the \texttt{core-game}. Thus, the LLM is judged to have made correct decisions only if it demonstrates forward-looking reasoning by choosing a sub-optimal solution in the \texttt{pre-game} to achieve the optimal solution in the \texttt{core-game}. We demonstrate a template to generate an nested form game in Appendix~\ref{appsec:additional:nested_form_game}.

Theoretically, using these atomic games, we can expand the framework to generate infinitely many increasingly complex game forms, thereby providing a continuous benchmark for evaluating the performance of more advanced LLMs.

\subsection{Evaluation Metrics}\label{sec:eval}
As explained in Section~\ref{sec:gt}, our benchmark is perfectly suitable to display in a 12x12 square table, each grid representing one of the 144 equivalence classes.
In the evaluation process we conduct \textit{repetitive} tests in every data point of each equivalence class. Each test starts with the input of the setting (classic/story-based) and the question, and ends with LLM's response containing a list of choices corresponding to multiple choices or no choice (when the given list is empty).

\textbf{Notation}.
For notation, we assign $\text{Freq}_{i, j, o}$ as the frequency of the $o$-th choice happening to be in the tests of the grid at $i$-th row, $j$-th column, where the 1, 2, 3 and 4-th choice correspond to the upper-left, upper-right, lower-left and lower-right quarter-grid respectively.

\textbf{Inconsistency Heat Map}.
According to conclusions of the Robinson-Goforth topology~\citep{robinson2005topology}, we convert the standard answer of each equivalence class into a heat map named the \textit{standard heat map}, with the coloured quarter-grid to be the choice in the standard answer. Similarly, as for practical result provided by LLMs, we set the value of $\text{Freq}_{i, j, o}$ as the colour depth of each quarter grid, which builds up the \textit{practical heat map}.
Naturally, we subtract the standard heat map from the practical heat map in an element-wise manner to get the \textit{inconsistency heat map}, which is a standardised tool for our evaluation, shown in Figure~\ref{fig:benchmark:consistency_heat_map}.

\textbf{Inconsistency Degree}.
In order to display the quantified performance of LLMs, we extract inconsistency degree from a map, which helps reveal the gap between LLMs' response and standard answer, and it is defined as $\mathrm{ID} = \frac{1}{144} \sum_{i=1}^{12} \sum_{j=1}^{12} \frac{1}{4} \sum_{o=1}^{4} \Delta \text{Freq}_{i, j, o} ^ 2$, where $\Delta \text{Freq}_{i, j, o}$ indicates the the difference (between the LLM's answer and the standard answer) of frequency of the $o$-th choice at $i$-th row, $j$-th column.


\textbf{Bias Degree}.
Owing to the symmetric property of the topology framework of 2×2 matrix games, the distribution of answers over the heat map has axial symmetry by the counter-diagonal (Figure~\ref{fig:benchmark:bias_score}). Motivated by this elegant property, we set up another metric to evaluate the bias degree of LLMs' answers, which we expect robuster LLMs to display lower degrees of bias.
The bias degree reflects the stability and symmetry of LLMs' strategy, and it is defined as
$\mathrm{BD} = \frac{1}{144} \sum_{i=1}^{12} \sum_{j=1}^{12} \frac{1}{4} \sum_{o=1}^4
(\text{Freq}_{i, j, o}-\text{Freq}_{j, i, \text{ref}_o}) ^ 2$,
where the meaning of $\text{ref}_o$ is the index of choice $o$'s counterpart considering the reflection operation by the counter-diagonal, and we have the mapping relation: $\{1, 2, 3, 4\} \mapsto \{4, 2, 3, 1\}$. (e.g. $\text{ref}_1=4$ means that the reflection counterpart of choice 1 is choice 4, vice versa)



\textbf{Perfect Accuracy Rate}.
In addition to the metrics mentioned above, we also set up a more rigorous metric named perfect accuracy rate, which ignores the partially correct answer and only considers perfectly correct answer in each test, and it is defined as $\mathrm{PAR} = 
\frac{1}{144} \sum_{i=1}^{12} \sum_{j=1}^{12} \frac{1}{T} \sum_{t=1}^{T} \mathbb{I}\{\text{rsp}_{t, i, j} = \text{std}_{i, j}\}$, which means that we count only if the response perfectly matches the standard answer, where $T$ represents the number of times we invoke a LLM to response on a certain game task.

\textbf{Metrics with Subscript}.
As a matter of fact, within the topology, different equivalence classes have different number of Nash equilibria (ranging from $\{0, 1, 2\}$), leading to a discrepancy in reasoning difficulty, therefore we propose metrics with subscript that represents for different types of equivalence groups (we refer them to 0-task, 1-task, 2-task respectively), which we refer to as sub-metrics. Therefore we have $\mathrm{ID}_n, \mathrm{BD}_n, \mathrm{PAR}_n (n = 0, 1, 2)$ which means the inconsistency degree, the bias degree, and the perfect accuracy rate across all equivalence classes that have $n$ equilibra.

%% file: sections/analysis.tex
\section{Analysis}\label{sec:exp}

\subsection{Overview of LLMs' Performance}
Overall, we select several SOTA models according to Open LLM Leaderboard~\citep{open-llm-leaderboard-v2} and conduct extensive experiments on {\bench}. These models include GPT (o3-mini-\texttt{2025-01-31}, gpt-4o-\texttt{2024-05-13}, gpt-4o-mini\texttt{-2024-07-18}, gpt-3.5-turbo\texttt{-0125}), Claude (claude-3-5-sonnet\texttt{-20240620}, claude-3-haiku\texttt{-20240307}), Llama (Llama-3.1-8B, Llama-3.1-70B), Gemma (gemma-3-27b-it), Qwen (Qwen3-32B, Qwen2-72B) and Deepseek (deepseek-reasoner). 
We perform 4 independent tests on each data point, covering both the classic setting and the story-based setting (thus we conduct 2,880 tests to evaluate a certain model). During the evaluation, we set the generation temperature to 0 or near 0, ensuring the lowest degree of uncertainty and enhancing the faithfulness of evaluation. More details of the evaluation process are provided in Appendix~\ref{appsec:dataset:pipeline}.

\input{tables/table_overall}

\begin{figure}[t]
    \centering
    \includegraphics[width=0.85\linewidth]{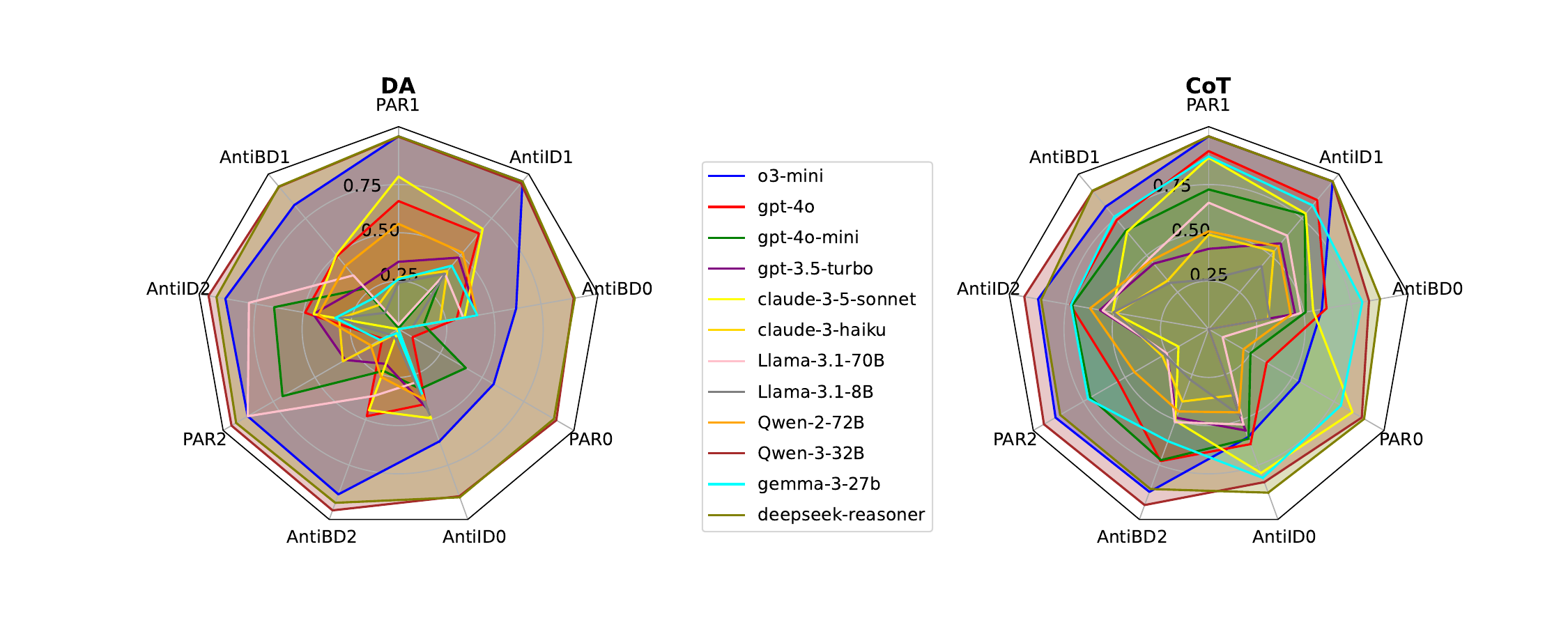}
    \caption{Radar charts of the 9 sub-metrics of LLMs' performance, comparing the DA prompting (left side) and the CoT prompting (right side). $\mathrm{AntiID}$ and $\mathrm{AntiBD}$ are derived from $\mathrm{ID}$ and $\mathrm{BD}$ while higher values indicate better performances (in order to consistent with $\mathrm{PAR}$).\protect\footnotemark}
    \label{fig:analysis:basic_result}
\end{figure}
\footnotetext{$\mathrm{AntiBD} = 1 - \sqrt \mathrm{BD}$, $\mathrm{AntiID} = 1 - \sqrt \mathrm{ID}$}

\begin{figure}[t]
\centering
     \begin{subfigure}[b]{0.405\textwidth}
         \centering
         \hspace*{-1.2cm}
         \includegraphics[width=\textwidth]{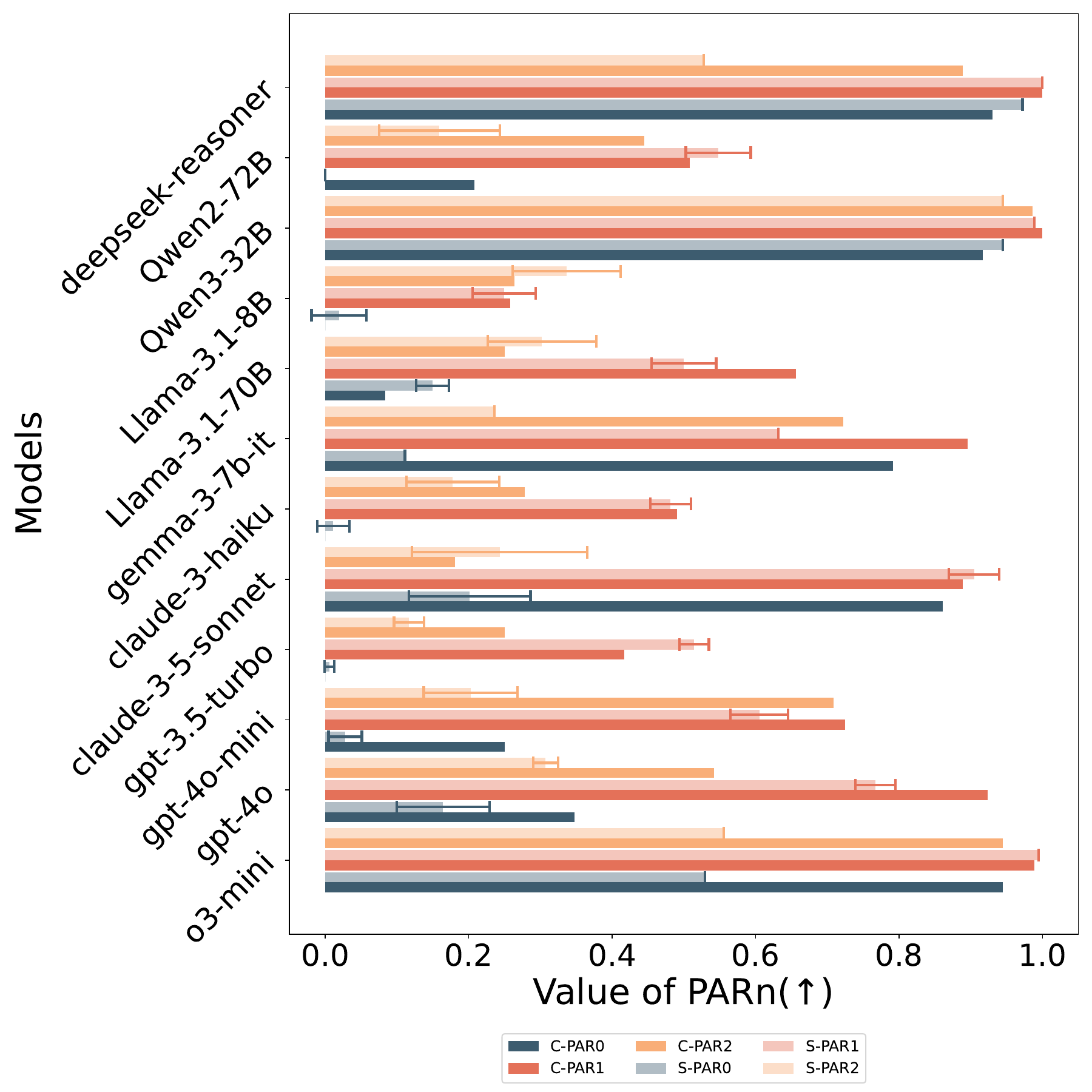}
         \caption{$\mathrm{PAR}_n$(↑)}
         \label{fig:analysis:PAR_by_settings}
     \end{subfigure}
     \begin{subfigure}[b]{0.27\textwidth}
         \centering
         \hspace*{-0.4cm}
         \includegraphics[width=\textwidth]{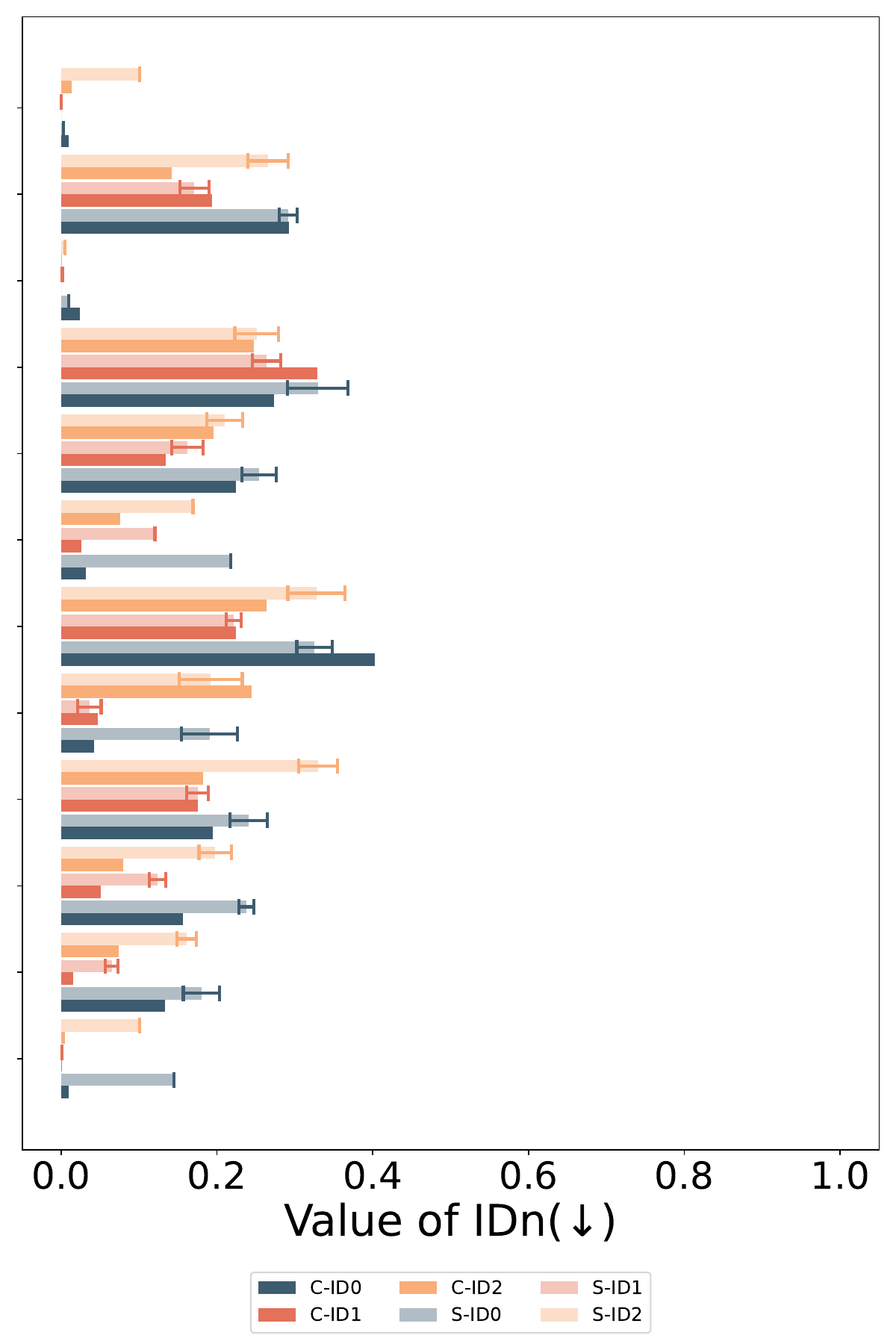}
         \caption{$\mathrm{ID}_n$(↓)}
         \label{fig:analysis:ID_by_settings}
     \end{subfigure}
     \begin{subfigure}[b]{0.27\textwidth}
         \centering
         \hspace*{0.2cm}
         \includegraphics[width=\textwidth]{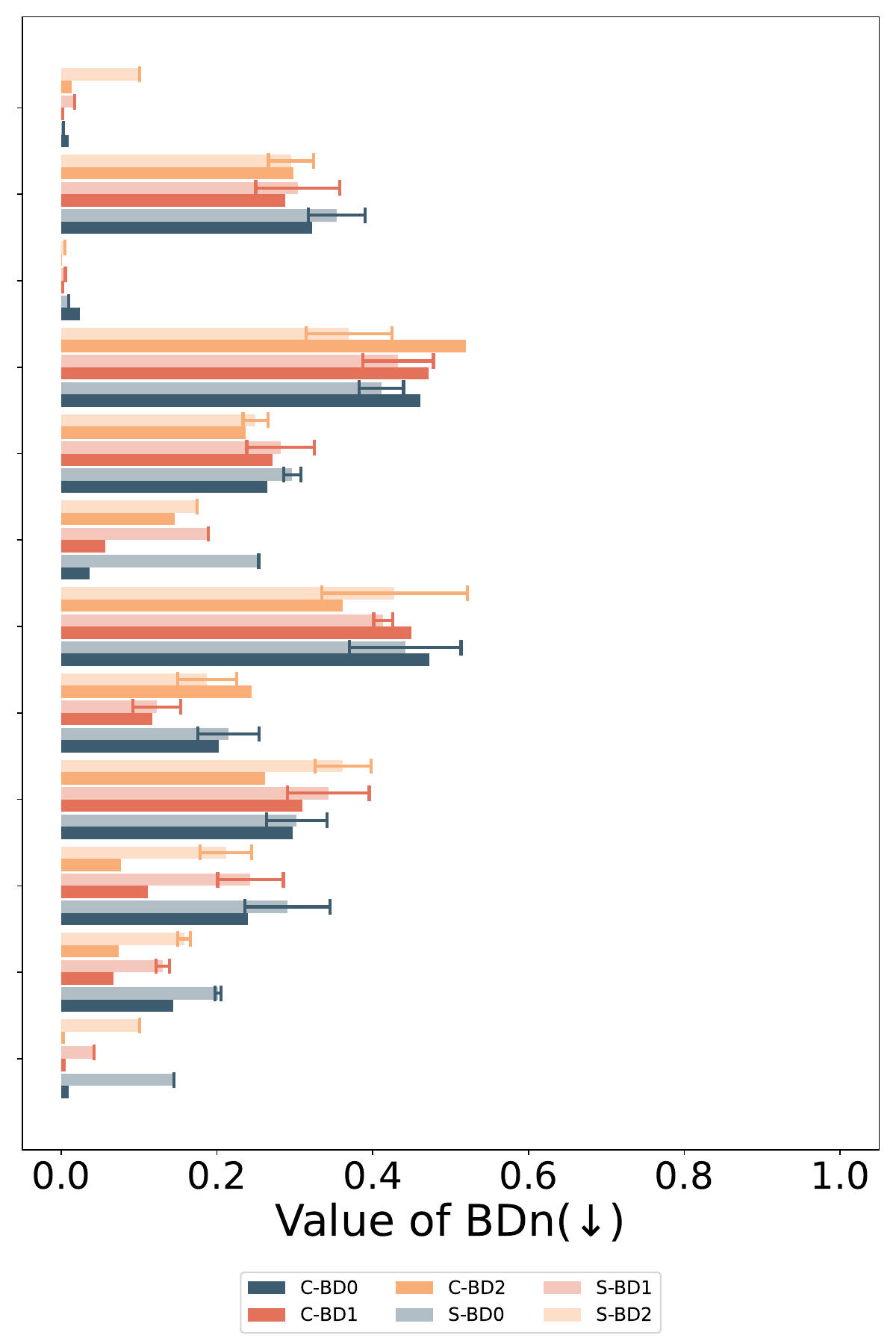}
         \caption{$\mathrm{BD}_n$(↓)}
         \label{fig:analysis:BD_by_settings}
     \end{subfigure}
\caption{Comparison of LLMs' performance under the classic setting (indicated by `\textit{C-}' label, in opaque colour) and the story-based setting (indicated by `\textit{S-}' label, in semi-opaque colour with error bar), where the length of the bars represent the value of metrics, and the error bars represent the standard deviation over all 5 data points of the story-based setting tasks.}
\label{fig:analysis:by_settings}
\end{figure}

\textbf{Games in {\bench} are not easy for most LLMs.}
First we overall evaluate how well LLMs can behave on the classic setting tasks of our benchmark, to assess their basic capability of strategic reasoning.
We initially adopt two basic prompting methods: Direct Answer (DA) prompting and Chain-of-Thought~ (CoT, \citep{wei2022chain}) prompting, which represent shallower, faster thinking patterns and deeper, slower thinking patterns, respectively.

As seen from Table~\ref{tab:basic_result_classic}, o3-mini, Qwen3-32B, deepseek-reasoner are more capable compared to other models, with a high overall accuracy rate (over 90\%) and low inconsistency and low bias score (around 1\%). 
Specifically, as shown in Figure~\ref{fig:analysis:basic_result} formed by 9 sub-metrics, Qwen3-32B and deepseek-reasoner almost surpass any other models on every aspect. Moreover, comparing the performance of employing DA prompting and CoT prompting, we find that CoT prompting almost provides comprehensive improvement. Despite the excellent performance of Qwen3-32B and deepseek-reasoner, other models often do not exhibit robust performance across all 3 different types of tasks. This indicates that even classic setting tasks from {\bench} are challenging for most LLMs.

\textbf{LLMs' performance is vulnerable across various narratives.}
At the theoretical level, we consider classic setting tasks and story-based tasks to be fundamentally the same problems within the domain of game theory. However, this conclusion appears not transferable to LLMs at the practical level. For LLMs, the complexity and nuance of story-based tasks introduce unique challenges, where LLMs are required to be robust in understanding and reasoning concurrently.

In Figure~\ref{fig:analysis:by_settings}, we compare the performance of LLMs using CoT prompting, which is more robust according to previous analysis. The figure reveals the vulnerable performance of LLMs on tasks in story-based setting (corresponding to various narratives), marked by two primary characteristics:

(1) Some models like o3-mini, gpt-4o, claude-3-5-sonnet and gemma-3-7b-it, exhibit significant performance degradation. Notably, o3-mini, gpt-4o and gemma-3-7b-it demonstrates a broad under-performance, while claude-3-5-sonnet shows the largest drop in 0-task, with its $S$-$\mathrm{PAR}_0$ metric reduced to less than one-fourth of $C$-$\mathrm{PAR}_0$, and its $S$-$\mathrm{ID}_0$ metric exceeding four times that of $C$-$\mathrm{ID}_0$.

(2) The performance of certain localities exhibits significant fluctuations. A particularly notable degradation occurs in the $\mathrm{PAR}$ scores for 0-task and 2-task scenarios handled by claude-3-5-sonnet, where the coefficients of variation $c_v$ (defined as $c_v = \frac{\sigma}{\mu}$, with $\sigma$ representing the standard deviation and $\mu$ the mean) approach 0.5. These eminent values of $c_v$ suggest a lack of robustness in performance across different narratives.


\input{tables/table_prompt}

\subsection{Findings of LLMs' Behaviours}
\textbf{LLMs demonstrate first/second-order ToM abilities.}
In all equivalence classes of tasks, 1-tasks have the lowest reasoning difficulty since at least one player has a dominant strategy, which means the player can make an unconditionally optimal decision regardless of the counterpart's choice. In such cases, once a player (denoted as A) can make unconditionally optimal decision, their counterpart (B) can, using first-order Theory-of-Mind (ToM), easily determine the best response for themselves (B).

This insight motivated us to apply FoToM prompting to LLMs, representing the \textbf{F}irst-\textbf{o}rder \textbf{T}heory-\textbf{o}f-\textbf{M}ind thinking, to aid in solving these tasks. As seen in Table~\ref{tab:comparing_prompt}, gpt-4o show improvement in both 0-tasks and 1-tasks when utilizing FoToM. Model claude-3-5-sonnet improves on 1-tasks and 2-tasks, and gpt-4o-mini displays a significant surge in performance on 1-tasks and so does Llama-3.1-70B on 2-tasks. However, no LLM achieves overall improvement across all task categories by merely using first-order ToM, and 0-tasks appear to be the most challenging for LLMs to solve.

Furthermore, we wondered if LLMs display some ability to use first-order ToM could also be capable of second-order ToM. According to \cite{liddle2006higher}, higher-order ToMs are generally more difficult to master than first-order ToM. Thus we selected only advanced models that demonstrated proficiency in first-order ToM to attempt solving specific tasks using \textbf{S}econd-\textbf{o}rder \textbf{T}heory-\textbf{o}f-\textbf{M}ind (SoToM) prompting. As seen in Table~\ref{tab:comparing_prompt}, gpt-4o, gpt-4o-mini and claude-3-5-sonnet show consistent performance when applying second-order ToM to tasks they are already capable of solving better with first-order ToM. However, the improvements from using SoToM generally do not exceed those achieved with first-order ToM. In addition, Llama-3.1-70B's underperformance with SoToM suggests that possessing first-order ToM capabilities does not necessarily imply proficiency with second-order ToM. The prompts used for FoToM and SoToM are provided in Appendix~\ref{appsec:dataset:prompt}.

\begin{figure}[t]
    \centering
    \begin{minipage}[b]{0.47\textwidth}
        \centering
        \includegraphics[width=\textwidth]{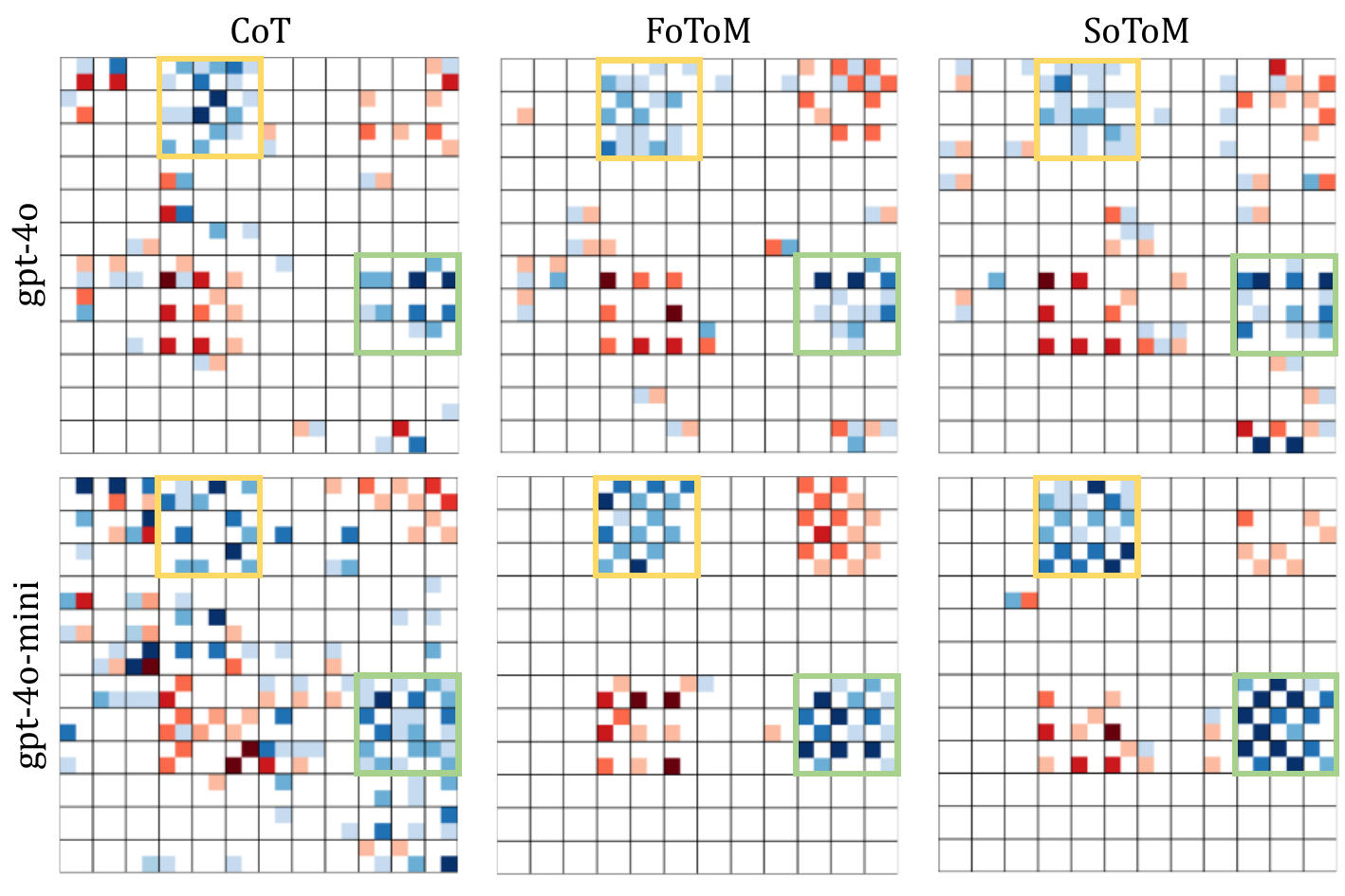}
        \caption{Inconsistency heat map of gpt-4o and gpt-4o-mini using different prompting methods. The yellow boxes and green boxes represent the 0-task areas in the topological framework.}
        \label{fig:analysis:scrutiny}
    \end{minipage}
    \hspace{0.5cm}
    \begin{minipage}[b]{0.475\textwidth}
        \centering
        \includegraphics[width=\textwidth]{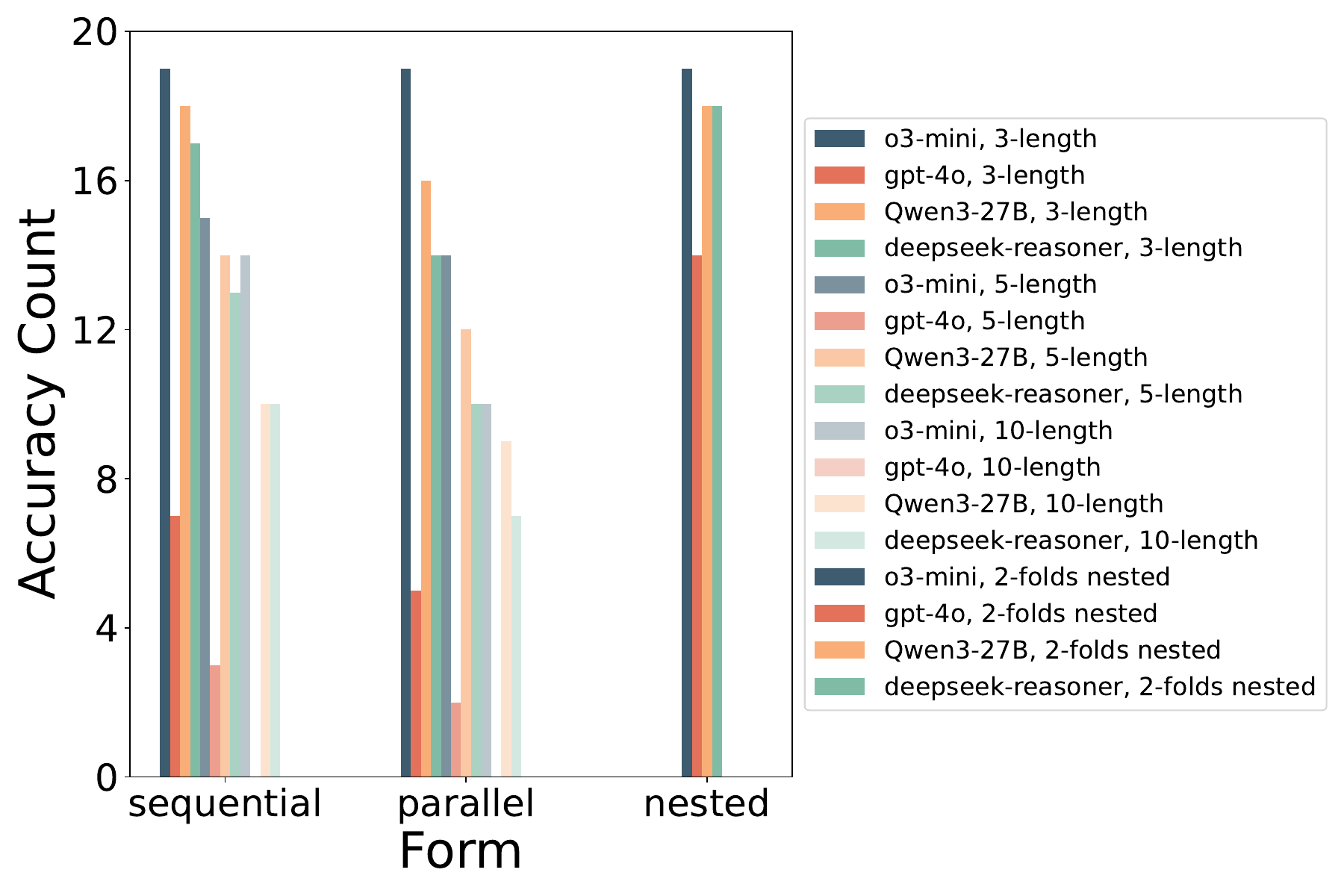}
        \caption{Advanced LLMs' performance on the games in complex forms of three types. Owing to the expensive inference cost, we run 20 times for each configuration.}
        \label{fig:performance_on_complex_forms}
    \end{minipage}
\end{figure}

\textbf{Certain behavioural pattern contributes to poor performance.}
Based on the analysis from the previous sections, it is encouraging to note that most LLMs demonstrate the best performance when solving 1-task scenarios, regardless of the prompting used (CoT, FoToM, or SoToM). However, some models, specifically gpt-4o and gpt-4o-mini, their performance declines significantly when addressing other types of tasks, and it is particularly noteworthy that they perform the worst on 0-tasks out of all types. Surprisingly, as illustrated in Figure~\ref{fig:analysis:scrutiny}, these models display a similar answering pattern that appears non-coincidental. Within the topological framework, there are two square areas representing 0-tasks (enclosed in \textcolor{goldenyellow}{yellow} boxes and \textcolor{grassgreen}{green} boxes), which should theoretically be symmetric across the counter-diagonal. The standard heat map of these two areas is entirely blank, reflecting no existing equilibrium, so the two areas of the inconsistency heat maps just reflect the distribution of LLMs' practical responses. Thus it is evident that the models exhibit a consistent pattern when addressing 0-tasks. In \textcolor{goldenyellow}{yellow}-box areas, their answers tend to emphasize the upper-right and lower-left quarter-grids, whereas in \textcolor{grassgreen}{green}-box areas, their answers tend to emphasize the upper-left and lower-right quarter-grids. This pattern appears to be the primary cause of the high bias degree.


To mitigate the possible effect of the position bias in the prompts (refer to the \textit{FoToM} prompting and \textit{SoToM} prompting in Appendix~\ref{appsec:dataset:prompt}), we design the \textit{reFoToM} prompting and the \textit{reSoToM} prompting (refer to the \textit{reFoToM} prompting and \textit{reSoToM} prompting in Appendix~\ref{appsec:dataset:prompt}) which swap the order of the players happens in the FoToM prompting and the SoToM prompting respectively. The results in Appendix~\ref{appsec:additional:asym_inconsistency} imply that such `asymmetric inconsistency pattern' is not strong related to the orders in the prompt. We demonstrate two typical examples of this phenomenon in Appendix~\ref{appsec:additional:failure_cases}.


\textbf{Complex forms bring more challenging tasks.}
To verify that {\bench} can be extended to harder tasks which may better align with complicated scenarios from the reality, we run the test on the three complex forms we mention in Section~\ref{sec:cxf}, to assess the performance of four strongest LLMs (o3-mini, gpt-4o, Qwen3-27B and deepseek-reasoner) in complex strategic reasoning.
We setup the test by dividing it into several types: (1) in sequential form and parallel form, we set the variable of number of the games from the set $\{3, 5, 10\}$; (2) in nested form, we just use some 2-folds nested games (due to the high verification cost when the number increases).

As seen from Figure~\ref{fig:performance_on_complex_forms}, among the four models, gpt-4o has a dramatically low accuracy rate in either sequential or parallel games. On the other hand, o3-mini performs best, but it still failed at times. When the number of the games increase, their performances both drop, which is consistent with intuition. As for the games of nested form, all models' performances are relatively reasonable, while it is fair to infer that if we increase the number of layers of the games that in the nested structures, it will present a great challenge for LLMs. 

%% file: tables/table_overall.tex
\begin{table}[t]
\centering
\small
\caption{Overall statistics of LLMs' performance on classic setting tasks. The up arrow(↑) means the larger value indicates better performance, while the down arrow(↓) means the smaller value indicates better performance. All values are expressed as percentages.
}

\resizebox{0.7\textwidth}{!}{
\begin{tabular}{@{}ccccccccc@{}}
\toprule
\multirow{4}{*}{\textbf{Family}} &
\multirow{4}{*}{\textbf{Model}} &
\multicolumn{6}{c}{\textbf{Metric / Prompting}} \\
\cmidrule(l){3-9} 
&
&
\multicolumn{2}{c}{\textbf{PAR(↑)}} &
\multicolumn{2}{c}{\textbf{ID(↓)}} &
\multicolumn{2}{c}{\textbf{BD(↓)}} \\
\cmidrule(l){3-9}
&
&
\textbf{DA} &
\textbf{CoT} &
\textbf{DA} &
\textbf{CoT} &
\textbf{DA} &
\textbf{CoT} \\

\midrule
\multirow{4}{*}{\textbf{GPT}} &
\textbf{o3-mini} &
{\color[HTML]{9A0000} \textbf{93.40}} &
{\color[HTML]{9A0000} \textbf{93.23}} &
{\color[HTML]{D88C8C} \textbf{1.91}} &
{\color[HTML]{D88C8C} \textbf{2.17}} &
{\color[HTML]{D88C8C} \textbf{3.82}} &
{\color[HTML]{D88C8C} \textbf{4.34}} \\
&
\textbf{gpt-4o} &
52.08 &
{\color[HTML]{D88C8C} \textbf{80.38}} &
16.81 &
{\color[HTML]{D88C8C} \textbf{3.78}} &
28.49 &
{\color[HTML]{D88C8C} \textbf{7.79}} &  \\
&
\textbf{gpt-4o-mini} &
14.93 &
74.02 &
27.15 &
4.38 &
48.59 &
8.29 \\
&
\textbf{gpt-3.5-turbo} &
30.21 &
34.38 &
27.64 &
17.87 &
50.15 &
30.19 & \\

\midrule
\multirow{2}{*}{\textbf{Claude}} &
\textbf{claude-3-5-sonnet} &
59.38 &
79.69 &
14.79 &
7.13 &
27.76 &
14.34 \\
&
\textbf{claude-3-haiku} &
24.31 &
40.28 &
39.58 &
25.17 &
72.22 &
44.10 \\
\midrule
\multirow{2}{*}{\textbf{Llama}} &
\textbf{Llama-3.1-70B} &
13.02 &
54.29 &
36.15 &
15.32 &
40.71 &
26.63 \\
&
\textbf{Llama-3.1-8B} &
18.75 &
22.63 &
38.49 &
31.19 &
81.32 &
47.64 \\
\midrule
\multirow{2}{*}{\textbf{Qwen}} &
\textbf{Qwen3-32B} &
{\color[HTML]{9A0000} \textbf{99.13}} &
{\color[HTML]{9A0000} \textbf{98.78}} &
{\color[HTML]{9A0000} \textbf{0.09}} &
{\color[HTML]{9A0000} \textbf{0.31}} &
{\color[HTML]{9A0000} \textbf{0.17}} &
{\color[HTML]{9A0000} \textbf{0.63}} & \\
&
\textbf{Qwen2-72B} &
43.06 &
46.21 &
26.30 &
19.94 &
35.59 &
29.29 \\
\midrule
\multirow{1}{*}{\textbf{Google}} &
\textbf{gemma-3-27b-it} &
20.83 &
{\color[HTML]{D88C8C} \textbf{86.11}} &
35.15 &
{\color[HTML]{D88C8C} \textbf{3.31}} &
63.35 &
{\color[HTML]{D88C8C} \textbf{6.55}} \\
\midrule
\multirow{1}{*}{\textbf{Deepseek}} &
\textbf{deepseek-reasoner} &
{\color[HTML]{9A0000} \textbf{98.78}} &
{\color[HTML]{9A0000} \textbf{97.74}} &
{\color[HTML]{9A0000} \textbf{0.09}} &
{\color[HTML]{9A0000} \textbf{0.29}} &
{\color[HTML]{9A0000} \textbf{0.17}} &
{\color[HTML]{9A0000} \textbf{0.59}} \\

\bottomrule
\end{tabular}}
\label{tab:basic_result_classic}
\end{table}

%% file: tables/table_prompt.tex
\begin{table}[t]
\centering
\small 
\caption{Performance of LLMs using different ToM compared to CoT. Text in red color indicates the performance gets better and text in blue color indicates the performance gets worse (both compared to CoT). Bold text means the best performance across the three prompting methods. Grey areas mean an LLM is good at using some kind(s) of ToM. All values are expressed as percentages.}
\resizebox{\textwidth}{!}{
\begin{tabular}{ccccccccccc}
\toprule
 &
   &
  \multicolumn{3}{c}{\textbf{0-Task}} &
  \multicolumn{3}{c}{\textbf{1-Task}} &
  \multicolumn{3}{c}{\textbf{2-Task}} \\ \cmidrule(l){3-11} 
\multirow{-3}{*}{\textbf{Model}} &
  \multirow{-3}{*}{\textbf{Prompting}} &
  \textbf{PAR\(_0\)(↑)} &
  \textbf{ID\(_0\)(↓)} &
  \textbf{BD\(_0\)(↓)} &
  \textbf{PAR\(_1\)(↑)} &
  \textbf{ID\(_1\)(↓)} &
  \textbf{BD\(_1\)(↓)} &
  \textbf{PAR\(_2\)(↑)} &
  \textbf{ID\(_2\)(↓)} &
  \textbf{BD\(_2\)(↓)} \\ \midrule
 &
  \textbf{CoT} &
  \cellcolor[HTML]{EFEFEF}34.72 &
  \cellcolor[HTML]{EFEFEF}13.37 &
  \cellcolor[HTML]{EFEFEF}14.41 &
  \cellcolor[HTML]{EFEFEF}92.36 &
  \cellcolor[HTML]{EFEFEF}1.58 &
  \cellcolor[HTML]{EFEFEF}6.76 &
  \textbf{54.17} &
  \textbf{7.38} &
  \textbf{7.38} \\
 &
  \textbf{FoToM} &
  \cellcolor[HTML]{EFEFEF}{\color[HTML]{CB0000} \textbf{43.06}} &
  \cellcolor[HTML]{EFEFEF}{\color[HTML]{CB0000} \textbf{9.46}} &
  \cellcolor[HTML]{EFEFEF}{\color[HTML]{CB0000} \textbf{9.81}} &
  \cellcolor[HTML]{EFEFEF}{\color[HTML]{CB0000} \textbf{95.14}} &
  \cellcolor[HTML]{EFEFEF}{\color[HTML]{CB0000} \textbf{0.72}} &
  \cellcolor[HTML]{EFEFEF}{\color[HTML]{CB0000} \textbf{4.14}} &
  {\color[HTML]{3166FF} 50.00} &
  {\color[HTML]{3166FF} 8.94} &
  {\color[HTML]{3166FF} 8.59} \\
\multirow{-3}{*}{\textbf{gpt-4o}} &
  \textbf{SoToM} &
  {\color[HTML]{3166FF} 31.94} &
  {\color[HTML]{CB0000} 9.81} &
  {\color[HTML]{CB0000} 10.68} &
  \cellcolor[HTML]{EFEFEF}{\color[HTML]{CB0000} 91.67} &
  \cellcolor[HTML]{EFEFEF}{\color[HTML]{CB0000} 1.45} &
  \cellcolor[HTML]{EFEFEF}{\color[HTML]{CB0000} 6.00} &
  {\color[HTML]{3166FF} 52.78} &
  {\color[HTML]{3166FF} 7.99} &
  {\color[HTML]{3166FF} 8.16} \\ \midrule
 &
  \textbf{CoT} &
  \textbf{25.00} &
  \textbf{15.62} &
  23.94 &
  \cellcolor[HTML]{EFEFEF}72.45 &
  \cellcolor[HTML]{EFEFEF}5.08 &
  \cellcolor[HTML]{EFEFEF}11.09 &
  \textbf{70.83} &
  7.97 &
  7.69 \\
 &
  \textbf{FoToM} &
  \textbf{25.00} &
  {\color[HTML]{3166FF} 19.53} &
  {\color[HTML]{CB0000} \textbf{19.53}} &
  \cellcolor[HTML]{EFEFEF}{\color[HTML]{CB0000} \textbf{99.54}} &
  \cellcolor[HTML]{EFEFEF}{\color[HTML]{CB0000} \textbf{0.03}} &
  \cellcolor[HTML]{EFEFEF}{\color[HTML]{CB0000} \textbf{5.08}} &
  {\color[HTML]{3166FF} 47.22} &
  {\color[HTML]{3166FF} 10.59} &
  {\color[HTML]{3166FF} 10.59} \\
\multirow{-3}{*}{\textbf{gpt-4o-mini}} &
  \textbf{SoToM} &
  {\color[HTML]{3166FF} 18.06} &
  {\color[HTML]{3166FF} 26.56} &
  {\color[HTML]{3166FF} 26.22} &
  \cellcolor[HTML]{EFEFEF}{\color[HTML]{CB0000} 98.84} &
  \cellcolor[HTML]{EFEFEF}{\color[HTML]{CB0000} 0.19} &
  \cellcolor[HTML]{EFEFEF}{\color[HTML]{CB0000} 5.38} &
  {\color[HTML]{3166FF} 68.06} &
  {\color[HTML]{CB0000} \textbf{5.38}} &
  {\color[HTML]{CB0000} \textbf{5.38}} \\ \midrule
 &
  \textbf{CoT} &
  \textbf{86.11} &
  \textbf{4.25} &
  20.23 &
  \cellcolor[HTML]{EFEFEF}88.89 &
  \cellcolor[HTML]{EFEFEF}4.72 &
  \cellcolor[HTML]{EFEFEF}11.68 &
  \cellcolor[HTML]{EFEFEF}18.06 &
  \cellcolor[HTML]{EFEFEF}24.48 &
  \cellcolor[HTML]{EFEFEF}24.48 \\
 &
  \textbf{FoToM} &
  {\color[HTML]{3166FF} 68.06} &
  {\color[HTML]{3166FF} 7.73} &
  {\color[HTML]{CB0000} \textbf{16.06}} &
  \cellcolor[HTML]{EFEFEF}{\color[HTML]{CB0000} \textbf{92.13}} &
  \cellcolor[HTML]{EFEFEF}{\color[HTML]{CB0000} \textbf{2.56}} &
  \cellcolor[HTML]{EFEFEF}{\color[HTML]{CB0000} \textbf{7.74}} &
  \cellcolor[HTML]{EFEFEF}{\color[HTML]{CB0000} \textbf{47.22}} &
  \cellcolor[HTML]{EFEFEF}{\color[HTML]{CB0000} 15.10} &
  \cellcolor[HTML]{EFEFEF}{\color[HTML]{CB0000} 15.10} \\
\multirow{-3}{*}{\textbf{claude-3-5-sonnet}} &
  \textbf{SoToM} &
  {\color[HTML]{3166FF} 47.22} &
  {\color[HTML]{3166FF} 21.35} &
  {\color[HTML]{3166FF} 28.99} &
  {\color[HTML]{CB0000} 90.05} &
  {\color[HTML]{CB0000} 4.05} &
  {\color[HTML]{3166FF} 14.38} &
  \cellcolor[HTML]{EFEFEF}{\color[HTML]{CB0000} 33.33} &
  \cellcolor[HTML]{EFEFEF}{\color[HTML]{CB0000} \textbf{14.93}} &
  \cellcolor[HTML]{EFEFEF}{\color[HTML]{CB0000} \textbf{14.93}} \\ \midrule
 &
  \textbf{CoT} &
  8.33 &
  22.47 &
  \textbf{26.43} &
  \textbf{65.59} &
  \textbf{13.43} &
  27.16 &
  \cellcolor[HTML]{EFEFEF}25.00 &
  \cellcolor[HTML]{EFEFEF}19.53 &
  \cellcolor[HTML]{EFEFEF}23.70 \\
 &
  \textbf{FoToM} &
  {\color[HTML]{3166FF} 2.78} &
  {\color[HTML]{3166FF} 30.82} &
  {\color[HTML]{3166FF} 35.59} &
  {\color[HTML]{3166FF} 49.54} &
  {\color[HTML]{3166FF} 18.68} &
  {\color[HTML]{3166FF} 27.49} &
  \cellcolor[HTML]{EFEFEF}{\color[HTML]{CB0000} \textbf{69.44}} &
  \cellcolor[HTML]{EFEFEF}{\color[HTML]{CB0000} \textbf{6.08}} &
  \cellcolor[HTML]{EFEFEF}{\color[HTML]{CB0000} \textbf{22.74}} \\
\multirow{-3}{*}{\textbf{Llama-3.1-70B}} &
  \textbf{SoToM} &
  {\color[HTML]{CB0000} \textbf{23.61}} &
  {\color[HTML]{CB0000} \textbf{21.27}} &
  {\color[HTML]{3166FF} 28.73} &
  {\color[HTML]{3166FF} 60.42} &
  {\color[HTML]{3166FF} 14.09} &
  {\color[HTML]{CB0000} \textbf{23.70}} &
  {\color[HTML]{3166FF} 12.50} &
  {\color[HTML]{3166FF} 24.05} &
  {\color[HTML]{3166FF} 25.26} \\ \bottomrule
\end{tabular}}
\label{tab:comparing_prompt}
\end{table}

%% file: sections/related_work.tex
\section{Related Work}

Large language models have made notable breakthroughs in reasoning tasks, such as mathematical, causal, and commonsense reasoning, enabling their increasing use in complex tasks that support human decision-making~\citep{imani2023mathprompter,kiciman2023causal,zhao2024large}. 
This progress has sparked a growing interest in studying their strategic reasoning capabilities~\citep{zhang2024llm}.
Game theory, with its highly abstract representation of real-world strategic scenarios, has garnered significant attention from researchers~\citep{duan2024gtbench,huang2024far}.
The prisoner's dilemma, as one of the most classical games, has been widely used to evaluate the strategic reasoning abilities of LLMs~\citep{brookins2023playing,guo2023gpt,akata2023playing,phelps2023investigating,xu2023magic}. 
In addition, several well-known game theory scenarios, such as the Dictator Game~\citep{horton2023large,Fan2023CanLL,brookins2023playing}, the Ultimatum Game~\citep{Aher2022UsingLL}, and the Public Goods Game~\citep{li2023beyond}, have been employed to evaluate LLMs' capabilities. 
However, current studies often focus on individual games, resulting in incomplete assessments and conclusions. 
To address this, we propose {\bench} for evaluating LLMs by 2×2 games, where its atomic games can be further organized using sequential, parallel, and nested formats to provide an in-depth evaluation of the SOTA models like o3-mini.

%% file: sections/discussion.tex
\section{Conlusion}

In this work, we introduce {\bench}, a benchmark for systematically evaluating the strategic reasoning abilities of LLMs by 2x2 matrix games. Based on Robinson-Goforth topology, we develop the classic setting tasks, and introduce various narratives based on story contexts generated by GPT-4o. By utilizing {\bench}, we can identify current flaws in LLMs' performance on these tasks, such as low accuracy rates and unstable inconsistency and bias degrees, even though the task difficulty is relatively moderate compared to many others. Additionally, when employing prompts to elicit their Theory-of-Mind thinkings on these tasks, some LLMs show improved performance, indicating that LLMs can, to some extent, master ToM and apply it in their reasoning processes. However, possessing first-order ToM abilities does not necessarily mean that LLMs will excel at mastering higher-order ToM. Furthermore, based on {\bench}, we introduce more forms of complex strategic reasoning tasks and pose a new challenge for LLMs.

%% file: sections/appendix.tex
\section{Brief Introduction to Strategic Reasoning}

\textbf{Definition.} Strategic reasoning~\citep{gandhi2023strategic,zhang2024llm} is a unique and sophisticated form of reasoning that focuses on making optimal decisions in multi-agent environments. It involves carefully selecting strategies by \textit{anticipating the actions of others} and \textit{understanding how one’s choices will influence their responses}. 

\textbf{Distinction.} What sets strategic reasoning apart is its \textit{dynamic nature} and the \textit{inherent uncertainty of adversarial actions}. Unlike other reasoning paradigms (commen sense reasoning, symbolic reasoning, casual reasoning, etc.), it demands a deep comprehension of ever-changing contexts and the ability to make rational, forward-thinking decisions based on the anticipated behaviors of others.

\textbf{Example.} In online advertising auctions~\citep{edelman2007internet}, advertisers compete for advertisement placements by bidding on specific audiences or keywords. Success depends on strategic reasoning, such as allocating budgets effectively, predicting competitors’ bids, and targeting audiences where competition is lower. Advertisers must also optimize their advertisement quality to reduce costs while maintaining visibility. Since auctions are dynamic and often follow a second-price model (where the winner pays just above the second-highest bid), advertisers continuously adjust their strategies to balance cost and competitiveness. This interplay of decisions makes advertising auctions a prime example of strategic reasoning in real-world applications. Considering scenarios where strategic reasoning can be applied with LLMs, fields such as societal simulation, economic simulation, game theory, and gaming~\citep{zhang2024llm} are prominent areas that often require this capability.

\textbf{Significance.} Strategic reasoning is a cornerstone for enabling intelligent systems to operate effectively in complex, multi-agent environments. In the context of LLMs, equipping them with strategic reasoning capabilities extends their potential beyond static information retrieval or pattern recognition tasks. It allows LLMs to simulate realistic decision-making processes, navigate dynamic social or economic systems, and collaborate or compete with other agents. This is particularly crucial in applications such as policy design, automated negotiations, and multi-agent simulations, where understanding and anticipating others’ behavior is essential for success. By fostering LLMs with strategic reasoning, we are able to bridge the gap between artificial intelligence and human-like adaptive decision-making, paving the way for more socially aware, context-sensitive, and intelligent systems that can tackle real-world challenges with greater precision and impact.

\section{Basic Things about Game Theory}
In this section, we discuss two fundamental concepts in game theory: dominant strategy and Nash equilibrium.

A dominant strategy is one that always provides a player with a payoff at least as high as any other strategy, regardless of the actions of other players. 
In other words, if a player has a dominant strategy, they will consistently choose it, as it either maximizes their payoff or does not reduce it, irrespective of the strategies chosen by others.

Nash equilibrium refers to a set of strategies, one for each player, where no player can benefit by unilaterally changing their strategy. 
At a Nash equilibrium, each player's strategy is the best response to the strategies of the other players. 
This means that if all players are following their Nash equilibrium strategies, no one has an incentive to deviate from their current strategy. 
It represents a stable state in the game where players' strategies are mutually optimal.

In many games, the dominant strategy equilibrium and Nash equilibrium may coincide, but not always. 
A dominant strategy equilibrium is a specific type of Nash equilibrium where each player has a strategy that is optimal regardless of others' strategies. 
However, in many cases, dominant strategies may not exist, requiring Nash equilibria to be identified through analysis and computation.

\section{Theory-of-Mind of LLMs}
Theory-of-Mind (ToM) refers to the ability to understand and infer human mental states~\citep{premack1988does}. 
Due to the multi-player nature of game theory, players' ability to reason about the ``minds" of other participants is crucial. 
Existing research has initiated discussions on whether machines possess ToM capabilities. 
For instance,~\citet{kosinski2023theory} suggested that ToM might emerge spontaneously in LLMs, as demonstrated through assessments using false-belief tasks. 
However,~\citep{ullman2023large} argued that such successes are fragile, easily disrupted by minor perturbations that would not affect an entity genuinely possessing ToM.
Nevertheless, many researchers propose enhancing LLMs' strategic reasoning abilities by incorporating ToM. 
\citet{guo2023suspicion} designed the Suspicion-Agent, which integrates a ToM-aware planning approach that leverages higher-order ToM capabilities, considering not only what the opponent might do (first-order ToM) but also what the opponent believes the Suspicion-Agent will do (second-order ToM). 
Additionally,~\citet{yim2024evaluating} introduced a ToM planning method in the Guandan poker game,~\citet{liu2024interintent} proposed an intention-guided mechanism,~\citet{xu2023magic} developed Probabilistic Graphical Modeling, and~\citet{zhang2024k} introduced K-Level-Reasoning, all utilizing ToM to enhance LLMs' strategic reasoning.
Given the broad application of ToM, this paper leverages {\bench} to comprehensively evaluate LLMs' ability to employ first-order and second-order ToM reasoning techniques for strategic reasoning.

\section{2×2 Matrix Game}
\subsection{Definition}\label{appsec:game:definition}
A normal-form game, commonly referred to as a 2×2 matrix game when involving two players each with two strategies, is a fundamental concept in game theory for representing strategic interactions. 
In this form, the game is depicted as a matrix, clearly outlining the players' strategies and corresponding payoffs. 
A typical 2×2 matrix game is structured as shown in Table~\ref{tab:2×2matrixgame}.

\begin{figure}[t]
     \centering
     \begin{subfigure}[b]{0.45\textwidth}
         \centering
         \includegraphics[width=0.8\textwidth]{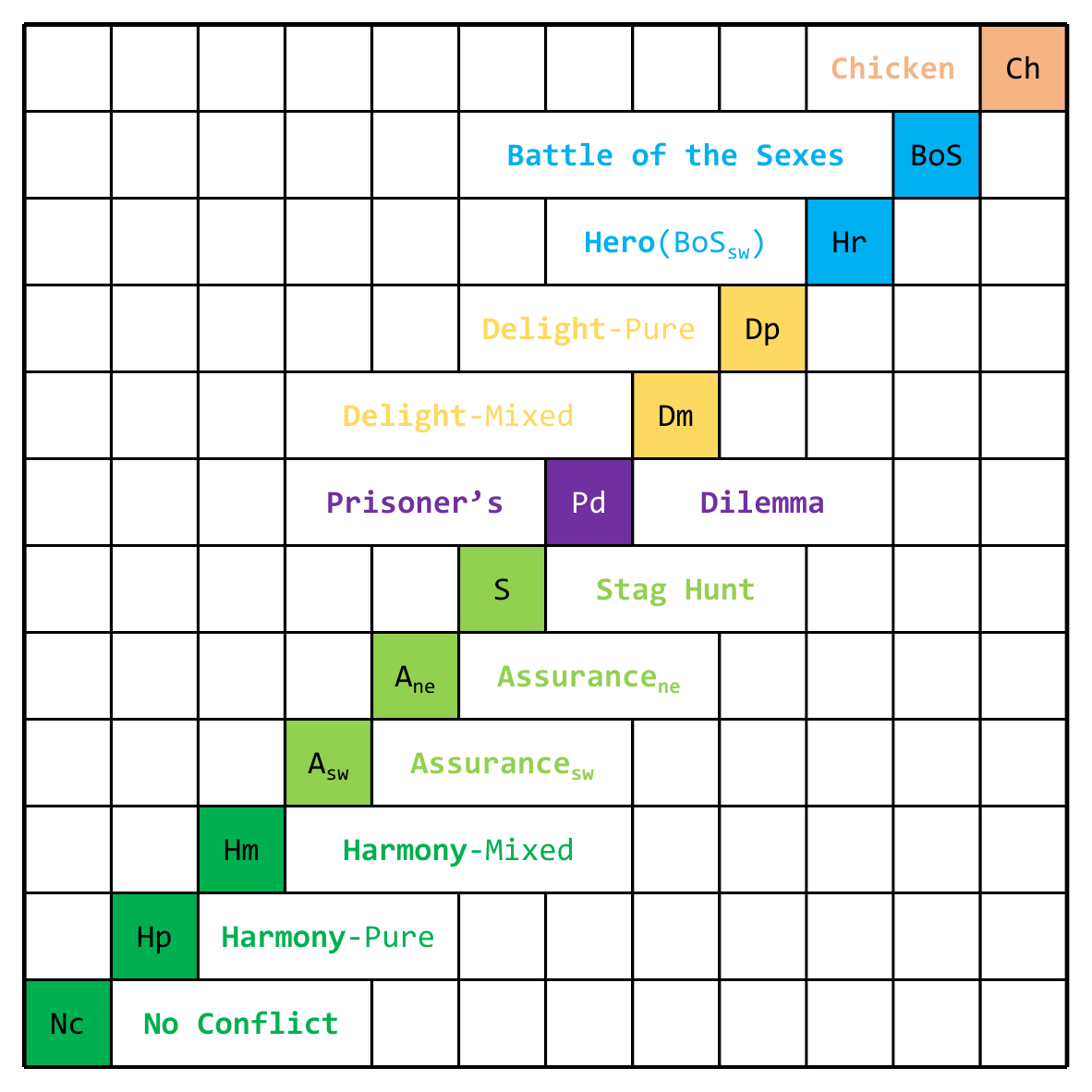}
         \caption{Most Famous Games}
         \label{fig:appendix:topology_famous}
     \end{subfigure}
     ~
     \begin{subfigure}[b]{0.45\textwidth}
         \centering
         \includegraphics[width=0.8\textwidth]{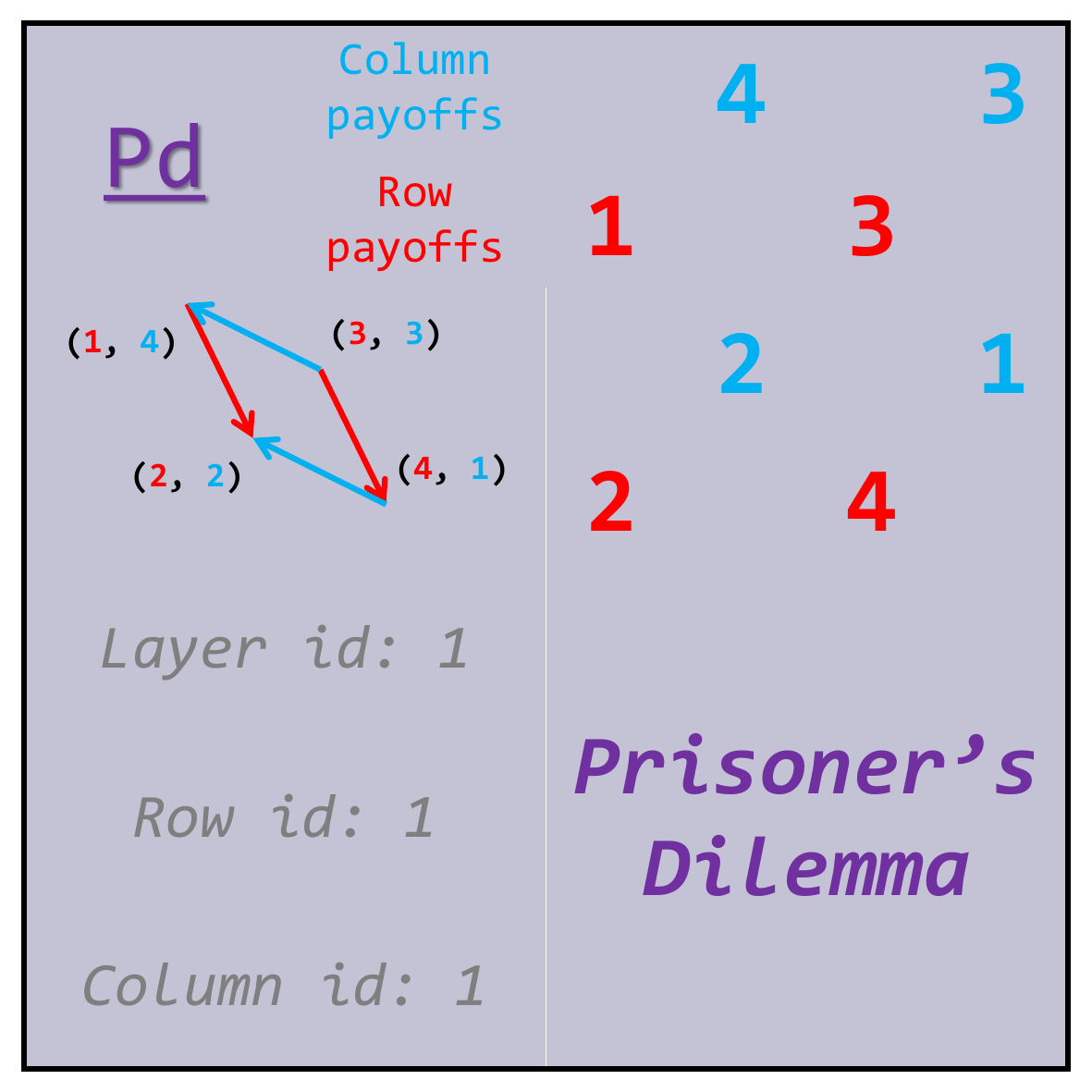}
         \caption{Details in a Grid}
         \label{fig:appendix:grid_details}
     \end{subfigure}
     \caption{The topology of the normal-form game system, which is presented by a square consisting of 12×12 grids. Figure~\ref{fig:appendix:topology_famous} displays the position of the most famous games in the topology. In each grid, there are specific details of the game, which is shown in Figure~\ref{fig:appendix:grid_details}.}
\label{fig:appendix:topology}
\end{figure}

\begin{table}[h]
\centering
\caption{The form of typical 2×2 matrix games.}
\begin{tabular}{|c|c|c|}
\hline
 & Player B: Strategy 1 & Player B: Strategy 2 \\
 \hline
Player A: Strategy 1 & (a, w) & (b, x) \\
\hline
Player A: Strategy 2 & (c, y) & (d, z)  \\
\hline
\end{tabular}
\label{tab:2×2matrixgame}
\end{table}

In this matrix, each cell represents the payoffs for both player A and player B, based on their chosen strategies. 
For instance, if player A selects strategy 1 and player B selects strategy 2, player A receives a payoff of \textit{a}, while player B receives a payoff of \textit{w}.

\subsection{Topology}
\label{appsec:game:topology}
Game theory research often concentrates on the Prisoner's Dilemma and a few other symmetric games, even though most potential games are asymmetric, and many ordinal games involve ties.

The findings on the topology of ordinal normal-form games~\citep{robinson2005topology} provide an elegant framework for systematically studying these games, encompassing all equivalence classes in an ordinal sense (where ``ordinal" refers to the ranking of payoffs rather than their specific values).

In this topological framework, as depicted in Figure~\ref{fig:appendix:topology}, well-known games such as the Prisoner's Dilemma, Stag Hunt, Battle of the Sexes, and Chicken are all symmetric and situated on the counter-diagonal of a 12×12 grid. 
The remaining games are located in the other grids, each with a corresponding ``sister game" that can be derived by reflecting across the counter-diagonal. 
A pair of sister games are identical when the roles of the two players are reversed.

Within each grid, basic information about the games in the equivalence classes is provided, including the family name and abbreviation, the payoff matrix, and the order graph, which illustrates the incentives for the row/column player to unilaterally change their choice for a higher payoff.

These 144 equivalence classes include 18 games with no equilibrium, 18 games with exactly two equilibria, and 108 games with a single equilibrium. 
Their distribution within the topology is symmetric across the counter-diagonal.

 \begin{figure}[h]
     \centering
     \includegraphics[width=0.36\textwidth]{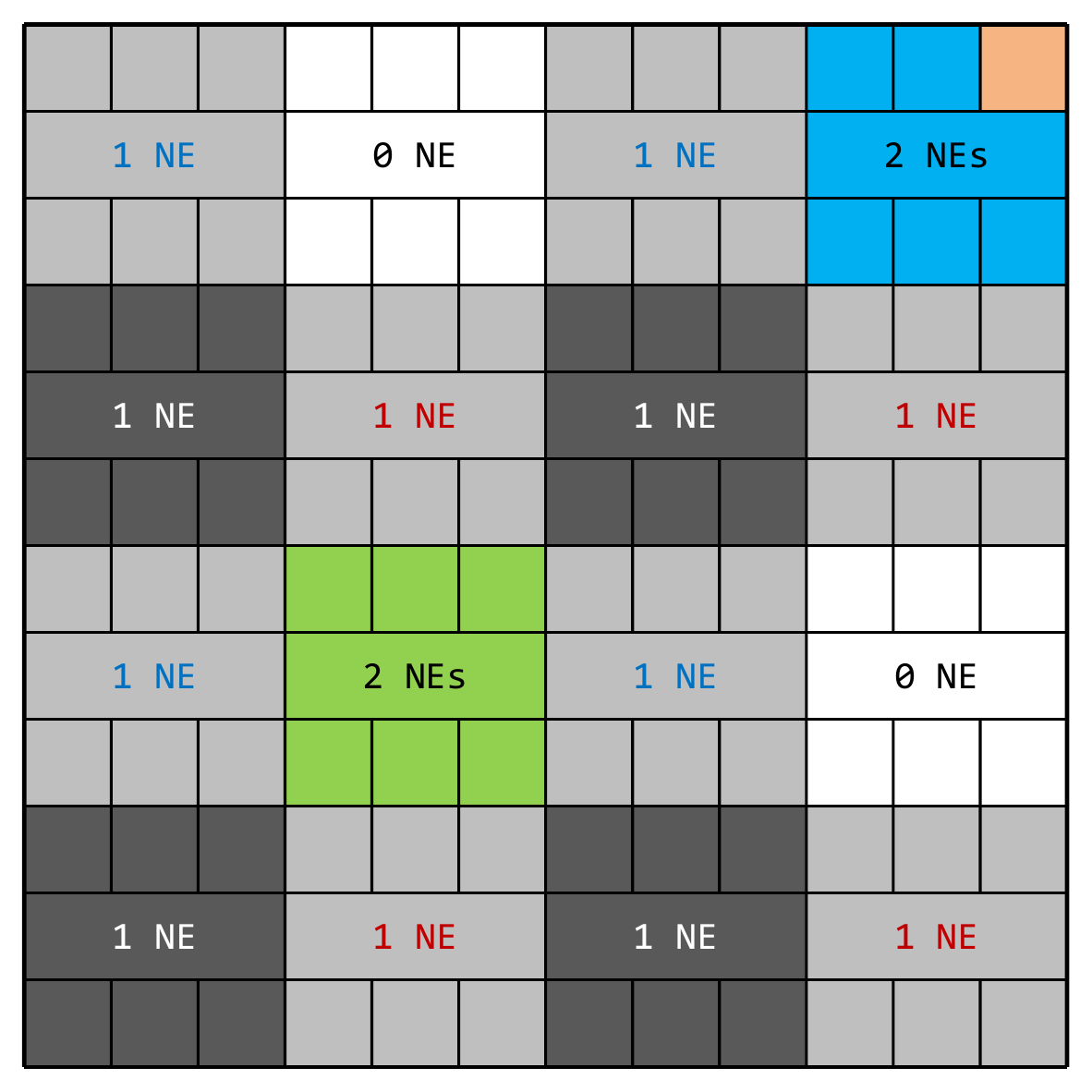}
     \caption{The distribution of games with 0, 1, or 2 Nash equilibria (a) is depicted according to the topology. Grids in grey indicate games with only 1 Nash equilibrium, while white grids represent games with no Nash equilibrium. Grids in other colours represent games with exactly 2 Nash equilibria. Text in blue/red indicates that the column/row player has a dominant strategy in the game, while white text signifies that both players have dominant strategies. In contrast, black text indicates that neither player has a dominant strategy.}
     \label{fig:appendix:topology_ne}
 \end{figure}

\subsection{Solution Structure}
As previously mentioned, all games in the topological framework can be categorized into three distinct groups based on the number of Nash equilibria. 
If we consider Nash equilibrium as the solution to finding stable strategy combinations, Figure~\ref{fig:appendix:topology_ne} illustrates the structure of these solutions.

In games with exactly one Nash equilibrium, at least one player (either the column player, row player, or both) has a dominant strategy, meaning they do not need to consider the other player's choice. 
These games are represented by grey or black grids.

Conversely, games with either 0 or 2 Nash equilibria share the characteristic that neither player has an unconditionally optimal choice, meaning no dominant strategies exist. 
However, in games with no Nash equilibrium (white grids), at least one player always has an incentive to unilaterally change their choice, regardless of the situation. 
In contrast, games with two Nash equilibria (orange, blue, or green grids) feature two stable strategy combinations.

Additionally, from a symmetry perspective, two sister games that are symmetric across the counter-diagonal belong to the same category and have identical Nash equilibria.

\section{More Information about our {\bench}}\label{appsec:dataset}
\subsection{Generation Pipeline}
\label{appsec:dataset:pipeline}
In our study, we design an efficient dataset generation pipeline that leverages GPT-4o as the core to produce the entire dataset, with rigorous human quality reviews incorporated. 
The pipeline is organized into three carefully designed stages:

\textbf{Classic Game Construction.}
Based on the topology of 2×2 games, we first introduce game descriptions for the payoff matrices of 144 game types, resulting in 144 classic games.
An example of a classic game is shown below, which mirrors the structure of the Prisoner's Dilemma. 
These 144 classic games will serve as seed games, with their inherent game structures generalized into more diverse, story-based games.

\begin{tcolorbox}[colframe=black, colback=white, boxrule=0.5mm, width=\textwidth, sharp corners=south, title=Example of classic game: \textit{classic/111}]
[Scenario]

Player A and Player B are playing a game. Either of them has two choices, namely A1, A2/B1, B2.
The payoff matrix of their different choice combinations is given below (larger number means higher
payoff):

\begin{Verbatim}
| A \ B | B1    | B2    |
|-------|-------|-------|
| A1    | 1 \ 4 | 3 \ 3 |
| A2    | 2 \ 2 | 4 \ 1 |
\end{Verbatim}

Both Player A and Player B are targeting maximizing their own payoff.

[/Scenario]
\end{tcolorbox}

\textbf{Story-based Game Generation.}
The aforementioned classic games offer a highly condensed mathematical representation of diverse game scenarios. 
However, in the real world, games often occur in complex social contexts involving various themes. 
To capture this complexity, we further designed \textit{story-based games}, incorporating richer entities and more intricate game scenarios.

Specifically, we used synthetic data generation techniques and crafted detailed prompts to set the construction constraints for generating high-quality story-based games. 
Additionally, to enhance the realism of our game scenarios, we manually defined several thematic categories to guide the data synthesis process (see \S\ref{appsec:dataset:diversity}). 
Both the prompt constraints and thematic categories ensure the generated content aligns with the intended structure and thematic elements.
An example of a generated story-based game is shown below, which follows the same game structure as the Prisoner's Dilemma and is presented within a new narrative context.
As such, the story-based game \texttt{story-based/111\_0} serves as a counterpart to the classic game \texttt{classic/111}.
For each classic game, we generate five corresponding story-based games. 
The data synthesis prompt is as follows. The red text are the placeholders for the variables of the generation code, where "domain" indicates the topic we random-choose for the task, and "matrix\_str" indicates the payoff matrix derived from the game structure we enumerate.

\begin{tcolorbox}[colframe=black, colback=white, boxrule=0.5mm, width=\textwidth, sharp corners=south, title=Story-based Game Generation Prompt]

Please generate a game theory short story with the following requirements:

- Specific topic: \textcolor{red}{\textit{\{domain\}}}

- There are two characters who may be in a situation of "cooperation" or "competition";

- Each character has 2 choices, and the combinations of their choices form 4 different scenarios;

- In these 4 scenarios, the two characters face different benefits/losses, which can be abstracted as different rewards they can obtain or different states they can achieve in each scenario;

- They each have a preference relationship for these rewards/states. We use numbers to represent the degree of preference, with 4 representing the most preferred and 1 the least preferred (i.e., preference degree 4\textgreater3\textgreater2\textgreater1);

- The payoff matrices for both characters can be abstracted and represented in one matrix, where A and B represent two characters and their choices are A1, A2/B1, B2. The respondent matrix is shown as below (the story you generate should have the same payoff structure as it):

\textcolor{red}{\textit{\{matrix\_str\}}}


Now please design a story that includes:

- Characters

- Each character's choices

- Characters' preferences for different scenarios

- Story description

Response format:

[Characters]

\{

\quad  ``A": ``...", ``B": ``..."

\}

[/Characters]

[Choices]

\{

\quad  ``A1": ``...", ``A2": ``...", ``B1": ``...", ``B2": ``..."

\}

[/Choices]

[Preferences]

Characters' preferences for different scenarios (4 most preferred, 1 least preferred):

\{

\quad  ``A": \{

\quad\quad    4: ``...", 3: ``...", 2: ``...", 1: ``..."

\quad   \},

\quad  ``B": \{

\quad\quad    4: ``...", 3: ``...", 2: ``...", 1: ``..."

\quad   \}

\}

[/Preferences]

[Payoff Matrix]

...

[/Payoff Matrix]

[Scenario]

...(to detailedly describe the situation, including the information of characters, choices, preferences and payoffs on different choice combinations)

[/Scenario]

\end{tcolorbox}

\begin{tcolorbox}[colframe=black, colback=white, boxrule=0.5mm, width=\textwidth, sharp corners=south, title=Example of story-based task: \textit{story-based/111\_0.txt}]
[Characters]

\{

\quad  ``A": ``Sarah, a small coffee shop owner",
  
\quad  ``B": ``Mark, owner of a large coffee chain"

\}

[/Characters]

[Choices]

\{

\quad  ``A1": ``Keep prices steady",

\quad  ``A2": ``Lower prices",

\quad  ``B1": ``Maintain current operations",

\quad  ``B2": ``Expand aggressively in Sarah's area"

\}

[/Choices]

[Preferences]

Characters' preferences for different scenarios (4 most preferred, 1 least preferred):

\{

\quad  ``A": \{

\quad\quad    4: ``Lower prices while Mark expands (survive and gain market share)",

\quad\quad    3: ``Keep prices steady while Mark maintains operations (stable coexistence)",

\quad\quad    2: ``Lower prices while Mark maintains operations (unnecessary price war)",

\quad\quad    1: ``Keep prices steady while Mark expands (lose customers)"

\quad   \},

\quad  ``B": \{

\quad\quad    4: ``Maintain operations while Sarah keeps prices steady (maximize profits)",

\quad\quad    3: ``Expand while Sarah keeps prices steady (gain market share)",

\quad\quad    2: ``Maintain operations while Sarah lowers prices (status quo)",

\quad\quad    1: ``Expand while Sarah lowers prices (costly competition)"

\quad   \}

\}

[/Preferences]

[Payoff Matrix]

\begin{Verbatim}
| A \ B | B1    | B2    |
|-------|-------|-------|
| A1    | 1 \ 4 | 3 \ 3 |
| A2    | 2 \ 2 | 4 \ 1 |
\end{Verbatim}

[/Payoff Matrix]

[Scenario]

In a bustling city, Sarah runs a cozy, independent coffee shop that has been a local favorite for years.

Recently, Mark's large coffee chain has been eyeing expansion into Sarah's neighborhood, threatening her business.
Sarah faces a critical decision: she can either keep her prices steady (A1) or lower them (A2) to attract more customers. Meanwhile, Mark must decide whether to maintain his current operations (B1)
or aggressively expand into Sarah's area (B2).
If Sarah keeps her prices steady and Mark maintains his current operations (A1, B1), Sarah struggles to compete (1) while Mark enjoys maximum profits (4). If Sarah lowers her prices and Mark stays put (A2, B1), both experience moderate success (2, 2) as Sarah attracts some new customers without
directly competing with Mark. Should Mark decide to expand while Sarah keeps prices steady (A1, B2), both would face challenges
but could coexist (3, 3) as Sarah retains loyal customers and Mark gains new ones. However, if Sarah lowers her prices as Mark expands (A2, B2), Sarah might survive and even gain market share (4), but Mark would face costly competition (1).

Their decisions will shape the local coffee market and determine the fate of Sarah's beloved shop.

[/Scenario]
\end{tcolorbox}

\textbf{Quality Verification.}
To ensure coherence and internal consistency in the generated games, we implement a multi-step generation strategy, incorporating meticulous human review.
First, GPT-4o generates an initial draft of the story, which is then reviewed by a human for any inconsistencies or logical flaws. 
If the draft fails this review, GPT-4o is prompted to identify the problematic sections and apply a self-correction mechanism.

During the self-correction phase, GPT-4o analyzes the story for inconsistencies and revises the flawed sections.
The revised version undergoes another round of human review. 
This iterative refinement process continues until the story meets the required quality standards.

If, after several rounds of regeneration, the story still contains significant issues or fails to meet the criteria, we may reject the output entirely. 
In such cases, the process is restarted from scratch with a new draft to ensure a fresh approach and to avoid perpetuating prior errors.

\subsection{Reasoning Prompt Used}
\label{appsec:dataset:prompt}
In this section, we present the prompts used by various reasoning methods. Notably, when invoking o1-mini to give response, we only use DA prompting, since the model are reported to perform reasoning internally and user should avoid`  prompting like chain-of-thought.

\begin{tcolorbox}[colframe=black, colback=white, boxrule=0.5mm, width=\textwidth, sharp corners=south, title=\textit{DA} prompting]
\textbf{System:} You are a spectator, and you should answer question based on given senario.

\textbf{User:}

\textit{\{task description\}}
\textit{\{task question\}}

Only give a block of python-style code containing your answer without any process. e.g.

```python

answer = [(``Ax", ``By")] \# list-type

'''
\end{tcolorbox}

\begin{tcolorbox}[colframe=black, colback=white, boxrule=0.5mm, width=\textwidth, sharp corners=south, title=\textit{CoT} prompting]
\textbf{System:} You are a spectator, and you should answer question based on given senario.

\textbf{User:}

\textit{\{task description\}}
\textit{\{task question\}}

Think step by step, and finally give a block of python-style code containing your answer. e.g.

```python

answer = [(``Ax", ``By")] \# list-type

'''
\end{tcolorbox}

\begin{tcolorbox}[colframe=black, colback=white, boxrule=0.5mm, width=\textwidth, sharp corners=south, title=\textit{FoToM} prompting]
\textbf{System:} You are a spectator, and you should answer question based on given senario.

\textbf{User:}

\textit{\{task description\}}
\textit{\{task question\}}

From A's perspective, try to \textcolor{orange}{figure out B's action} and make choice. Then from B's perspective try to \textcolor{orange}{figure out A's action} and make choice. Finally as a spectator, give a block of python-style code containing your answer. e.g. 

```python

answer = [(``Ax", ``By")] \# list-type

'''
\end{tcolorbox}

\begin{tcolorbox}[colframe=black, colback=white, boxrule=0.5mm, width=\textwidth, sharp corners=south, title=\textit{SoToM} prompting]
\textbf{System:} You are a spectator, and you should answer question based on given senario.

\textbf{User:}

\textit{\{task description\}}
\textit{\{task question\}}

From A's perspective, try to \textcolor{orange}{figure out B's action}, note that \textcolor{purple}{he may also reason based on your information or reasoning}. Then from B's perspective try to \textcolor{orange}{figure out A's action}, note that \textcolor{purple}{he may also reason based on your information or reasoning}. Finally as a spectator, give a block of python-style code containing your answer. e.g. 

```python

answer = [(``Ax", ``By")] \# list-type

'''
\end{tcolorbox}

\begin{tcolorbox}[colframe=black, colback=white, boxrule=0.5mm, width=\textwidth, sharp corners=south, title=\textit{reFoToM} prompting]
\textbf{System:} You are a spectator, and you should answer question based on given senario.

\textbf{User:}

\textit{\{task description\}}
\textit{\{task question\}}

From B's perspective, try to figure out A's action and make choice. Then from A's perspective try to figure out B's action and make choice. Finally as a spectator, give a block of python-style code containing your answer. e.g. 

```python

answer = [(``Ax", ``By")] \# list-type

'''
\end{tcolorbox}

\begin{tcolorbox}[colframe=black, colback=white, boxrule=0.5mm, width=\textwidth, sharp corners=south, title=\textit{reSoToM} prompting]
\textbf{System:} You are a spectator, and you should answer question based on given senario.

\textbf{User:}

\textit{\{task description\}}
\textit{\{task question\}}

From B's perspective, try to figure out A's action, note that he may also reason based on your information or reasoning. Then from A's perspective try to figure out B's action, note that he may also reason based on your information or reasoning. Finally as a spectator, give a block of python-style code containing your answer. e.g. 

```python

answer = [(``Ax", ``By")] \# list-type

'''
\end{tcolorbox}

\subsection{Benchmark Diversity}\label{appsec:dataset:diversity}
Our dataset is characterized by the diverse contexts encapsulated within the story-based tasks, a diversity that manifests across several dimensions.

\begin{figure}[!p]
     \centering
     \begin{subfigure}[b]{0.75\textwidth}
         \centering
         \includegraphics[width=\textwidth]{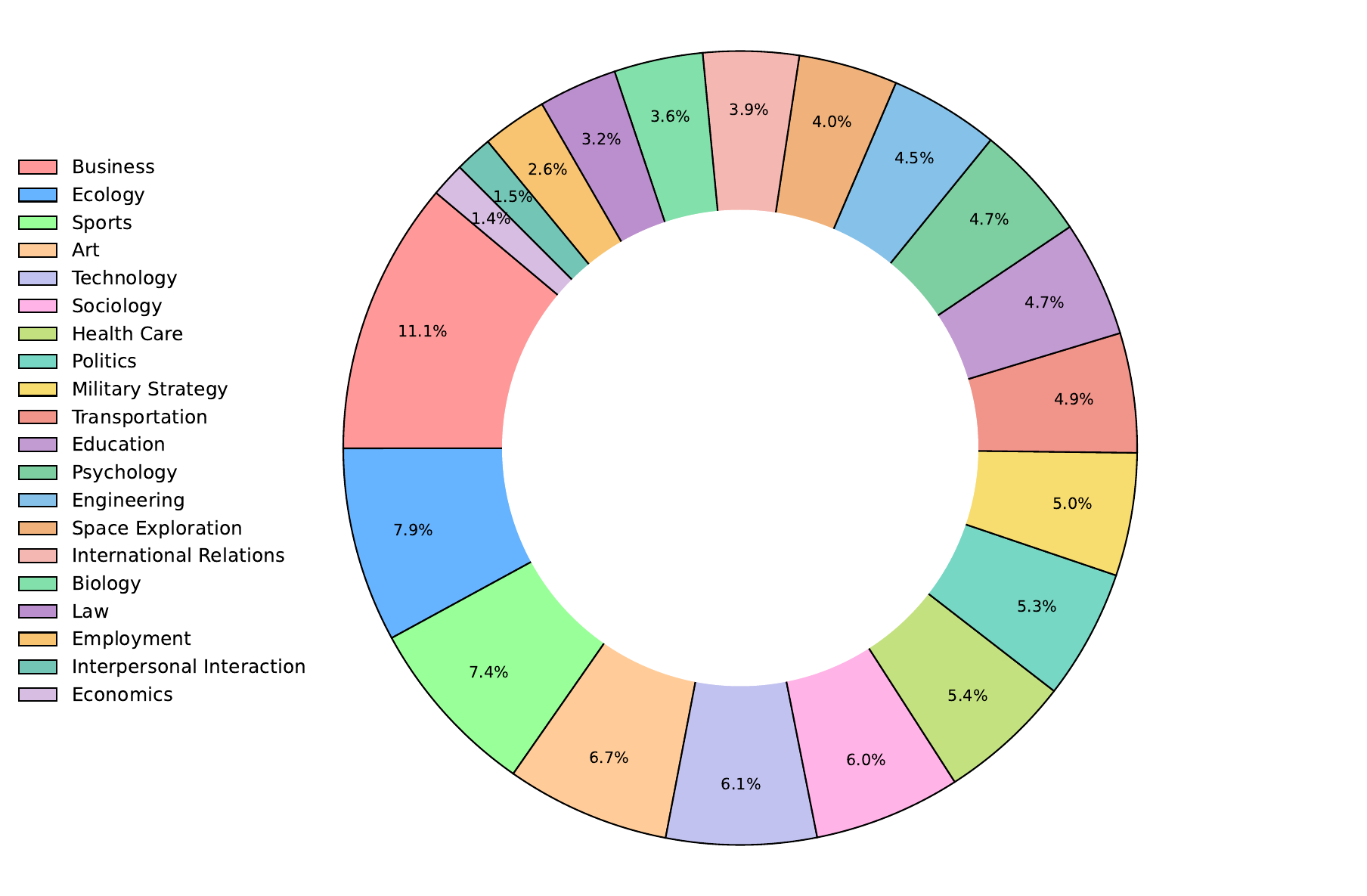}
         \caption{The topic distribution of story-based games.}
         \label{fig:appendix:topic_distribution}
     \end{subfigure}
     ~
     \begin{subfigure}[b]{0.75\textwidth}
         \centering
         \includegraphics[width=\textwidth]{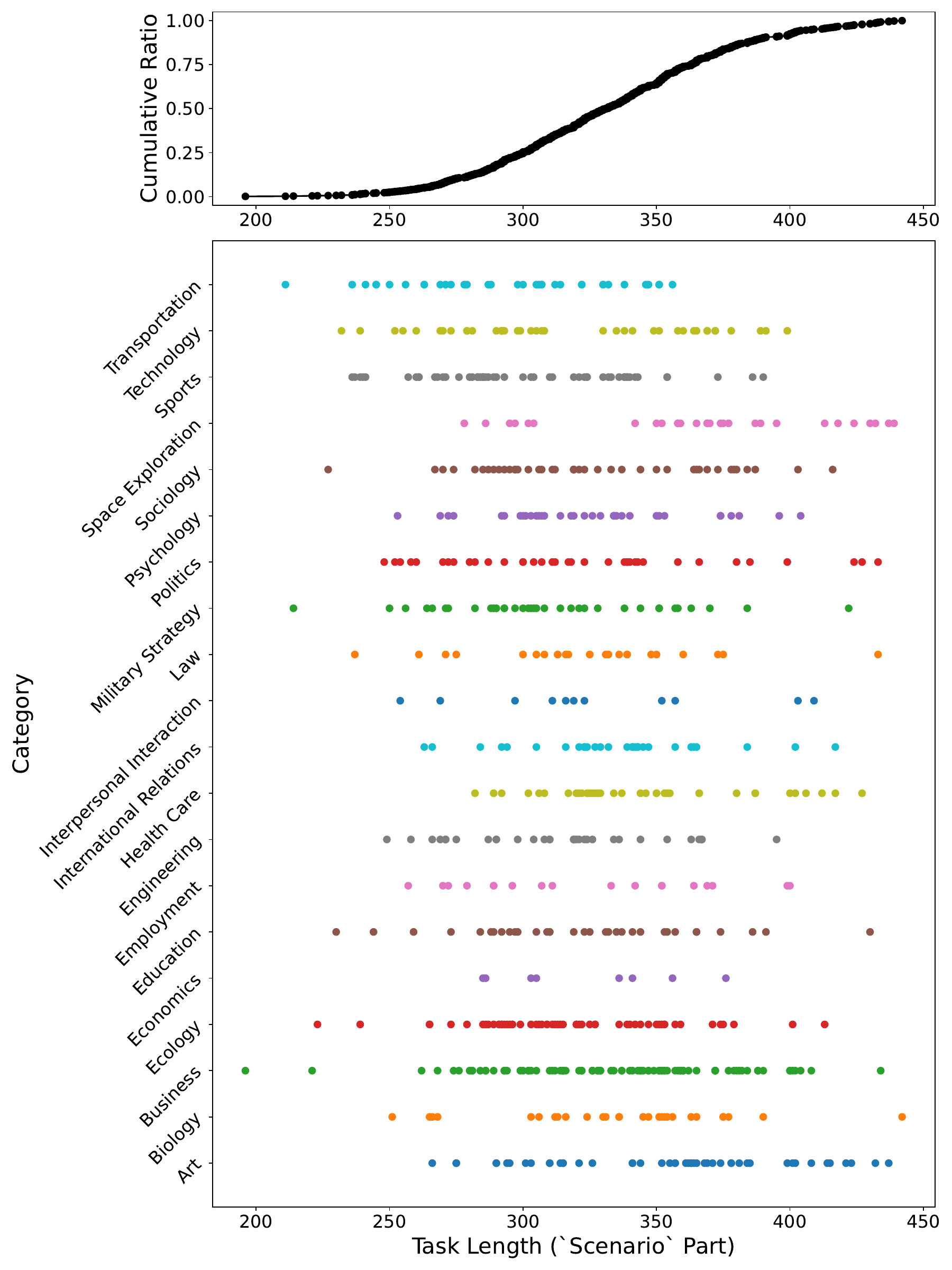}
         \caption{Cumulative distribution of lengths by ratio and scatter plot of lengths by topic.}
         \label{fig:appendix:length_distribution}
     \end{subfigure}
\caption{Statistical distribution of story-based games over 20 topics.}
\label{fig:appendix:dataset_diversity}
\end{figure}

Firstly, we have identified 20 distinct topics derived from everyday life scenarios where cooperation and competition are likely to occur. 
These topics align with situations commonly depicted in various game families. 
The distribution of story-based games across these 20 topics is visualized in Figure~\ref{fig:appendix:topic_distribution}.

The topics encompass a broad spectrum of fields, including Business, Ecology, Sports, Technology, Health Care, Politics, and more. 
Notably, Business constitutes the largest proportion of the dataset at 11.1\%, while the remaining topics are more evenly distributed, with percentages generally ranging from approximately 1.4\% to 7.9\%.

Given the nature of these long-text reasoning tasks, the scenarios within our story-based games typically range from 200 to 450 words in length. 
As illustrated in Figure~\ref{fig:appendix:length_distribution}, over 90\% of scenario lengths fall within the 250 to 400-word interval. 
Additionally, we provide a scatter plot of scenario lengths by topic to further demonstrate the diversity of our generated dataset.

\section{Additional Results}
\subsection{Asymmetric Inconsistency Pattern}
\label{appsec:additional:asym_inconsistency}
\begin{figure}[t]
    \centering
    \includegraphics[width=0.6\linewidth]{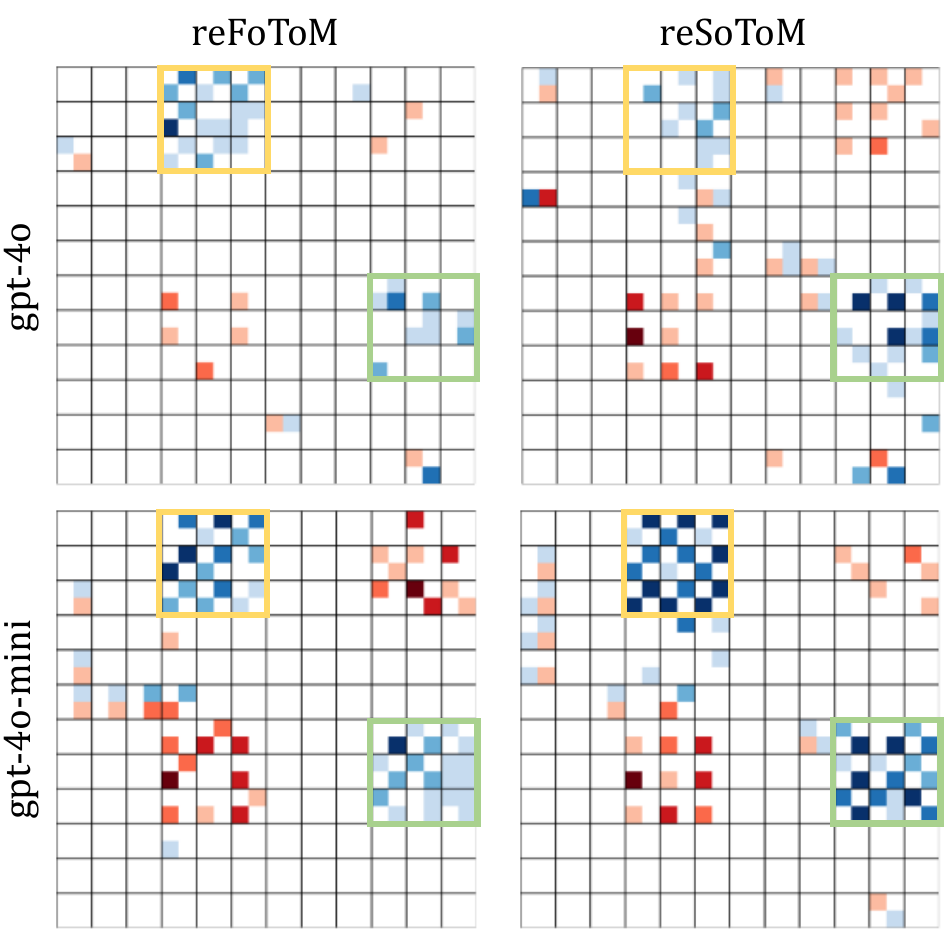}
    \caption{Inconsistency heat map of GPT series models using reFoToM and reSoToM prompting.}
    \label{fig:appendix:additional_scrutiny}
\end{figure}

We show in Figure~\ref{fig:appendix:additional_scrutiny} that GPT series models still display similar pattern when using reFoToM and reSoToM prompting. Yellow-box areas and green-box areas display an asymmetric inconsistency pattern.

In order to further quantify how significant does the results display such pattern, we also propose a metric named significance degree which confined in [0, 1] and it is defined as
\begin{align*}
\mathrm{SD} = 
\frac{1}{18} \sum_{i, j} \mathbb{I}\{\mathrm{\#NE}(i, j) = 0\} \cdot \frac{1}{4} (\text{Freq}_{i, j, 1} + \text{Freq}_{i, j, 4} - \text{Freq}_{i, j, 2} - \text{Freq}_{i, j, 3}) * \mathrm{S}(i, j)
\end{align*}
where $\mathrm{\#NE}(i, j)$ represents the number of Nash equilibria of the tasks of grid $(i, j)$, and $\mathrm{S}(i, j)$ is decided by the area that the grid $(i, j)$ belongs to, having value 1 given the grid in the green area and value -1 given the grid in the yellow area.

We present the statistical results of LLMs' performance in Table~\ref{tab:significance_degree}, which show that the $\mathrm{SD}$ values for using ReFoToM are similar to those for FoToM, and the values for ReSoToM are close to those for SoToM.

Additionally, the results indicate that employing ToM helps gpt-4o reduce the asymmetric inconsistency pattern, while it conversely makes gpt-4o-mini more 'stubborn' in maintaining this pattern. Furthermore, higher-order ToM appears to have a stronger effect than first-order ToM.

\begin{table}[]
\centering
\caption{The significance degree of top-tier GPT models performance. The larger value indicates the higher significance of the peculiar answering pattern. Near-zero value means no particular pattern. All values are expressed as percentages.}
\begin{tabular}{cccccccc}
\toprule
\textbf{Model} &
  \textbf{CoT} &
  \textbf{FoToM} &
  \textbf{ReFoToM} &
  \multicolumn{2}{c}{\cellcolor[HTML]{FFFFFF}\textbf{SoToM}} &
  \multicolumn{2}{c}{\cellcolor[HTML]{FFFFFF}\textbf{ReSoToM}} \\ \midrule
\textbf{gpt-4o} &
  13.89 &
  9.38 &
  8.33 &
  \multicolumn{2}{c}{\cellcolor[HTML]{FFFFFF}4.51} &
  \multicolumn{2}{c}{\cellcolor[HTML]{FFFFFF}6.25} \\ \midrule
\textbf{gpt-4o-mini} &
  5.56 &
  26.74 &
  20.49 &
  \multicolumn{2}{c}{\cellcolor[HTML]{FFFFFF}32.64} &
  \multicolumn{2}{c}{\cellcolor[HTML]{FFFFFF}35.42} \\ \bottomrule
\end{tabular}
\label{tab:significance_degree}
\end{table}

\subsection{Cases of Failure}
\label{appsec:additional:failure_cases}
In this section, we demonstrate two examples of the real response from gpt-4o-mini, which derived from a pair of symmetric data points (the grid (row 3, column 4) and the grid (row 9, column 10)) that theoretically have the same answer and similar reasoning pattern. We use the red coloured texts to indicate the mistaken parts of the reasoning processes.

\begin{tcolorbox}[colframe=black, colback=white, boxrule=0.5mm, width=\textwidth, sharp corners=south, title=row 3 column 4]
To determine the most likely choice combinations for Players A and B, we can analyze the payoff matrix step by step.

1. **Identify the Payoffs**: The payoff matrix is as follows:
\begin{Verbatim}
   | A \ B | B1    | B2    |
   |-------|-------|-------|
   | A1    | 3 \ 1 | 1 \ 4 |
   | A2    | 2 \ 3 | 4 \ 2 |
\end{Verbatim}

2. **Player A's Choices**:

\quad   - If Player B chooses B1:
   
\quad\quad     - A1 gives Player A a payoff of 3.
     
\quad\quad     - A2 gives Player A a payoff of 2.
   
\quad   - If Player B chooses B2:
   
\quad\quad     - A1 gives Player A a payoff of 1.
     
\quad\quad     - A2 gives Player A a payoff of 4.

\quad   From Player A's perspective:
   
\quad\quad   - Against B1, A1 (3) is better than A2 (2).
   
\quad\quad   - Against B2, A2 (4) is better than A1 (1).

3. **Player B's Choices**:

\quad   - If Player A chooses A1:
   
\quad\quad     - B1 gives Player B a payoff of 1.
     
\quad\quad     - B2 gives Player B a payoff of 4.

\quad   - If Player A chooses A2:
   
\quad\quad     - B1 gives Player B a payoff of 3.
     
\quad\quad     - B2 gives Player B a payoff of 2.

\quad   From Player B's perspective:
   
\quad\quad   - Against A1, B2 (4) is better than B1 (1).
   
\quad\quad   - Against A2, B1 (3) is better than B2 (2).

4. **Best Responses**:
   
\quad   - If Player A chooses A1, Player B will choose B2 (since 4 \textgreater 1).
   
\quad   - If Player A chooses A2, Player B will choose B1 (since 3 \textgreater 2).
   
\quad   - If Player B chooses B1, Player A will choose A1 (since 3 \textgreater 2).
   
\quad   - If Player B chooses B2, Player A will choose A2 (since 4 \textgreater 1).

\textcolor{red}{5. **Nash Equilibria**:}
   
\quad   \textcolor{red}{- The combination (A1, B2)} gives payoffs (3, 4).
   
\quad   \textcolor{red}{- The combination (A2, B1)} gives payoffs (2, 3).

6. **Conclusion**:
   
\quad   - The most likely choice combinations are (A1, B2) and (A2, B1). Since both combinations are viable and lead to different outcomes, we conclude that both are possible.    

Thus, the final answer in the required format is:

```python

answer = [(``A1'', ``B2''), (``A2'', ``B1'')] \# list-type

'''
\end{tcolorbox}

\begin{tcolorbox}[colframe=black, colback=white, boxrule=0.5mm, width=\textwidth, sharp corners=south, title=row 9 column 10]
To determine the most likely choice combinations for Players A and B, we can analyze the payoff matrix step by step.

1. **Identify the Payoffs**:

\quad   - For Player A:
   
\quad\quad     - If A chooses A1:
     
\quad\quad\quad       - Against B1: Payoff = 2
       
\quad\quad\quad       - Against B2: Payoff = 4
     
\quad\quad     - If A chooses A2:
     
\quad\quad\quad       - Against B1: Payoff = 3
       
\quad\quad\quad       - Against B2: Payoff = 1
   
\quad   - For Player B:
   
\quad\quad     - If B chooses B1:
     
\quad\quad\quad       - Against A1: Payoff = 4
       
\quad\quad\quad       - Against A2: Payoff = 2
     
\quad\quad     - If B chooses B2:
     
\quad\quad\quad       - Against A1: Payoff = 1
       
\quad\quad\quad       - Against A2: Payoff = 3

2. **Best Responses**:
   
\quad   - If Player B chooses B1:
   
\quad\quad     - Player A's best response is A2 (3 \textgreater 2).
   
\quad   - If Player B chooses B2:
   
\quad\quad     - Player A's best response is A1 (4 \textgreater 1).
   
\quad   - If Player A chooses A1:
   
\quad\quad     - Player B's best response is B1 (4 \textgreater 1).
   
\quad   - If Player A chooses A2:
   
\quad\quad     - Player B's best response is B2 (3 \textgreater 2).

\textcolor{red}{3. **Nash Equilibria**:}
   
\quad   - A Nash Equilibrium occurs when both players are choosing their best responses to each other.
   
\quad   - From the analysis:
   
\quad\quad     \textcolor{red}{- (A1, B1) is a Nash Equilibrium because A's best response to B1 is A2, but B's best response to A1 is B1.}
     
\quad\quad     \textcolor{red}{- (A2, B2) is also a Nash Equilibrium because A's best response to B2 is A1, but B's best response to A2 is B2.}

4. **Conclusion**:
   
\quad   - The most likely choice combinations are (A1, B1) and (A2, B2) since both are Nash Equilibria.

Now, we can represent this in the required Python-style code:

```python

answer = [(``A1'', ``B1''), (``A2'', ``B2'')] \# list-type

'''

\end{tcolorbox}
\newpage
\section{About Complex Form Game}
\subsection{Example of Different Complex Form}
\begin{figure}[h]
    \centering
    \includegraphics[width=0.8\linewidth]{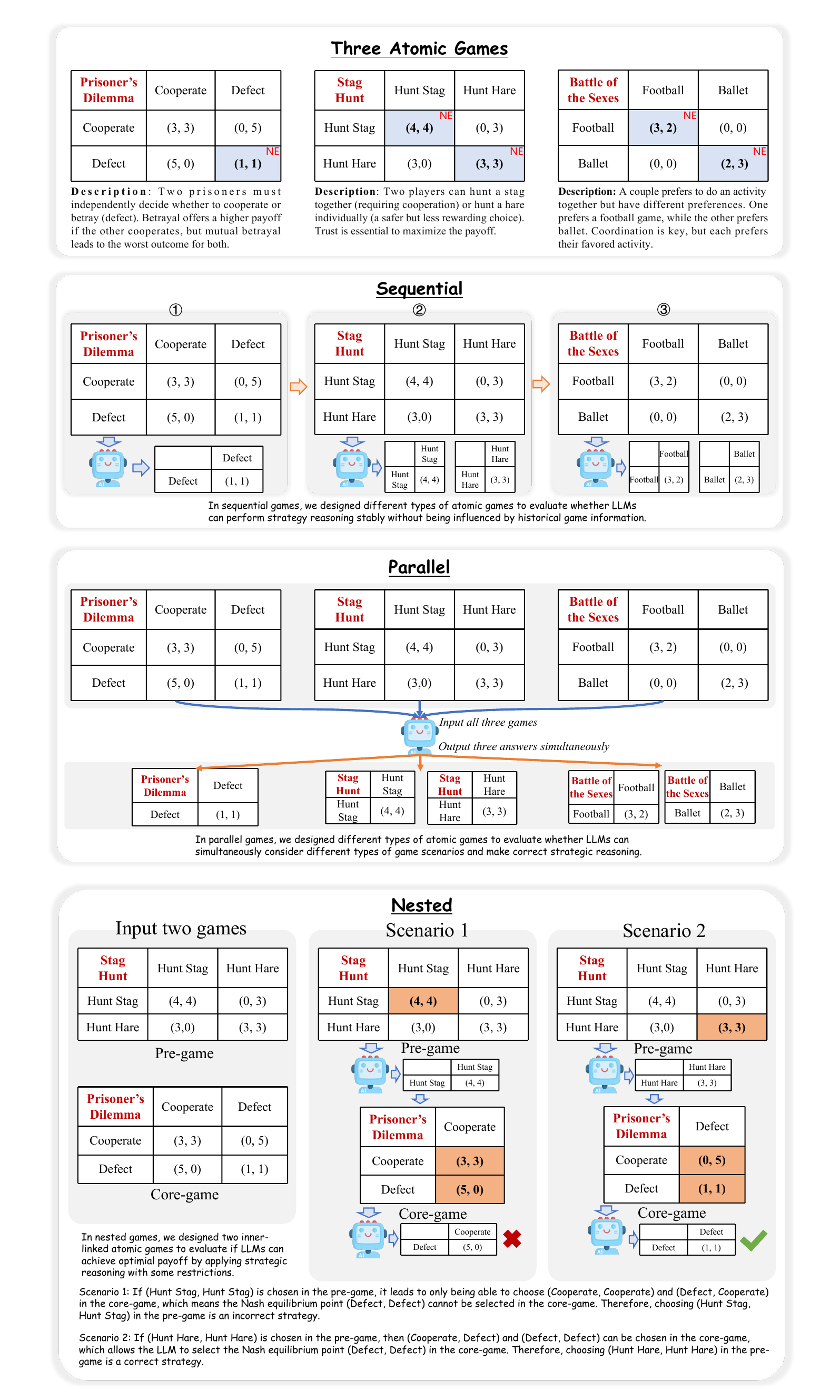}
    \caption{An example of how we build sequential, parallel, and nested game from some of the atomic games in {\bench}.}
    \label{fig:complex_example}
\end{figure}

In this section, we use visualized example (referred to Figure~\ref{fig:complex_example}) to illustrate different kinds of complex forms.

\subsection{Nested Form Game}
\label{appsec:additional:nested_form_game}
In this section, we demonstrate the template we use for generating nested form games. The text in red are the placeholders for the variables of the generation codes.

\begin{tcolorbox}[colframe=black, colback=white, boxrule=0.5mm, width=\textwidth, sharp corners=south, title=template of a kind of nested form game]

\textless Pre-Game \textgreater

\textcolor{red}{\{pre\_game\}}

\textless Core-Game \textgreater

\textcolor{red}{\{core\_game\}}

[Question]

Player A and B are facing the two games, the pre-game and the core-game.

Note that their final goal is to maximize own payoff first in the core Game, then in the pre-game.

Additionally, \textcolor{red}{\{restricted\_player\}} is attached with an restriction that if the situation of the pre-game is \textcolor{red}{\{restricted\_situation\}}, then he can not choose action \textcolor{red}{\{restricted\_choice\}}.

What is/are the most possible choice combination(s) of the pre-game ultimately? (when all choice combinations have equal possibility, the answer should contain nothing)

[/Question]
\end{tcolorbox}

After a nested form game is generated through our template, we still need to check if the Nash equilibria of the pre-game changes after the restriction from the core game. If the set of Nash equilibria does change, then we use this as a piece of data to evaluate LLMs, observing if they can observe such a violation of original NEs' structure.

\subsection{Significance of Atomic Games and Complex-Form Games}

Our evaluation of complex-form games serves as a test of whether LLMs can solve real-world problems with higher complexity, rather than merely solving a single atomic game in isolation. The atomic games in {\bench} represent the \textit{primary components} of complex real-world social scenarios~\citep{gintis2014bounds}. In addition to the typical combinations found in temporal contexts (i.e., sequential games) or spatial contexts (i.e., parallel games), which require reasoning and independent decision-making, \cite{tsebelis1990nested} introduced a concept known as \textit{nested games}, where two or more games are inner-linked. This type of game composition often arises in real-world domains such as politics and economics, where decisions made in one arena can influence or constrain decisions in another.

\section{Data Leakage Concern}
There have been concerns regarding potential data leakage in our synthetic dataset. However, we believe such concerns can largely be mitigated based on the following analyses.

\subsection{Source Analysis}
Our dataset is synthetic and template-based, which significantly reduces the likelihood of explicit contamination. Since the data are generated using predefined templates and rules, there is a very low probability that any real-world data or previously encountered examples could seep into the dataset. This template-based approach helps maintain consistency across the examples, ensuring that the LLMs are evaluated purely on their strategic reasoning capabilities rather than being influenced by previously seen examples.

\subsection{Perplexity (PPL) Analysis}
We also performed a perplexity analysis on the dataset, testing it with those open-source models. Our findings show that the PPL values are within reasonable ranges, indicating that the dataset does not exhibit typical signs of contamination:

For shorter classic data points, the PPL values range from 8 to 10.

\begin{table}
\centering
\caption{The average value and the standard deviation of the perplexity of the open-sourced models on the classic data points.}
\begin{tabular}{cccc}
\toprule
\textbf{Model} &
  \textbf{Avg} &
  \textbf{Std} \\ \midrule
\textbf{Llama-3.1-8B} &
  8.9532 &
  0.1733 \\ \midrule
\textbf{Llama-3.1-70B} &
  8.0626 &
  0.1552 \\ \midrule
\textbf{Qwen-72B} &
  9.0817 &
  0.0913 \\ \bottomrule
\end{tabular}
\label{}
\end{table}

For longer story-based data points, the PPL values range from 3 to 6.

\begin{table}[h]
\centering
\caption{The average value and the standard deviation of the perplexity of the open-sourced models on the story-based data points.}
\begin{tabular}{cccc}
\toprule
\textbf{Model} &
  \textbf{Avg} &
  \textbf{Std} \\ \midrule
\textbf{Llama-3.1-8B} &
  5.0239 &
  0.2308 \\ \midrule
\textbf{Llama-3.1-70B} &
  4.2513 &
  0.1921 \\ \midrule
\textbf{Qwen-72B} &
  3.8923 &
  0.1389 \\ \bottomrule
\end{tabular}
\label{}
\end{table}

These PPL values suggest that the data is not overly predictable, indicating that the dataset is not contaminated. The variation in perplexity across different models demonstrates that there is no significant bias or discrepancy in predicting any of the data points. This further suggests that there are no obvious signs of data leakage. However, while we have not observed any direct evidence of leakage, we acknowledge that we cannot fully rule out the possibility of subtle contamination.

\subsection{More Details about the Concern}
we would like to again emphasize that the conclusions drawn from our source and perplexity (PPL) analyses are not contradictory to the observations about famous vs. unfamous games: our {\bench} has a relatively low risk of data leakage.

First, the dataset of {\bench} features on the synthetic context and we use a standardized template to develop it, and even for the story-based part, they have such long context (200-450 tokens per data point, as shown in \ref{fig:appendix:length_distribution}). This substantially lowers the likelihood of any entire paragraph being exposed to LLMs during training.

Also, for the classic part, from our original experiments and additional ablations, it is possible that some games have been exposed to LLMs, but it will not be a risk, and we provide some reasoning as follow:

\begin{itemize}
    \item The normal level of perplexity tells us that there is low possibility of exposing an entire same data point of our {\bench} to LLMs, for the definition of perplexity is to measure to what extent LLM can correctly predict next token based on the preceding context.

    \item Some finding of the paper, along with the additional finding on famous vs. unfamous games, indicate that some LLMs might be familiar with some of the games, while this kind of familiarity, is not derived from the leakage of our data point. Instead, it is most probably originated from their emerging ability that they utilize to apply such familiar knowledge to do better in reasoning. For example, the training corpus of an LLM may include content which indirectly related to a famous game like \textit{The Prisoner’s Dilemma} rather than an unfamous games, so LLMs have a chance to know more information about the game with similar game structure. (This is similar to solve a maths problem or a card game, both need some pre-knowledge/experience which can boost ones' performance.)
    
    \item Actually, in many prior studies \citep{Aher2022UsingLL, horton2023large, guo2023suspicion, duan2024gtbench}, some classic games (not necessarily bi-matrix games, but well-known in game theory or economics) are still being employed to evaluate if LLMs can conduct strategic reasoning like humans or even perform better. Compared to these work, {\bench} have a lower risk of data leakage because our data points are much longer and we use a synthetic method which incorporates much more variables.
\end{itemize}
In conclusion, while famous games might be easier for LLMs due to their inherent familiarity, this familiarity is not a consequence of {\bench} data leakage. Instead, it reflects the models' ability to leverage prior knowledge, further demonstrating the value of TMGBench in assessing strategic reasoning.

\section{Limitations}
Our {\bench} focuses on a very specific area within the vast domain of game theory, highlighting the fact that there is still a significant portion of game theory that lacks systematic exploration. While it may be infeasible or even impossible to develop a framework that is suitable for all types of games, we hope that benchmarks like {\bench} can help identify issues such as inherent imbalances and the non-robustness of LLMs' strategic reasoning abilities.

%% file: neurips_2025.bbl
\begin{thebibliography}{47}
\providecommand{\natexlab}[1]{#1}
\providecommand{\url}[1]{\texttt{#1}}
\expandafter\ifx\csname urlstyle\endcsname\relax
  \providecommand{\doi}[1]{doi: #1}\else
  \providecommand{\doi}{doi: \begingroup \urlstyle{rm}\Url}\fi

\bibitem[Aher et~al.(2022)Aher, Arriaga, and Kalai]{Aher2022UsingLL}
Gati Aher, RosaI. Arriaga, and Adam~Tauman Kalai.
\newblock Using large language models to simulate multiple humans and replicate human subject studies.
\newblock In \emph{International Conference on Machine Learning}, 2022.
\newblock URL \url{https://api.semanticscholar.org/CorpusID:251719353}.

\bibitem[Akata et~al.(2023)Akata, Schulz, Coda-Forno, Oh, Bethge, and Schulz]{akata2023playing}
Elif Akata, Lion Schulz, Julian Coda-Forno, Seong~Joon Oh, Matthias Bethge, and Eric Schulz.
\newblock Playing repeated games with large language models.
\newblock \emph{ArXiv preprint}, abs/2305.16867, 2023.
\newblock URL \url{https://arxiv.org/abs/2305.16867}.

\bibitem[Brookins \& DeBacker(2023)Brookins and DeBacker]{brookins2023playing}
Philip Brookins and Jason~Matthew DeBacker.
\newblock Playing games with gpt: What can we learn about a large language model from canonical strategic games?
\newblock \emph{Available at SSRN 4493398}, 2023.

\bibitem[Carlsson \& Van~Damme(1993)Carlsson and Van~Damme]{carlsson199312}
Hans Carlsson and Eric Van~Damme.
\newblock 12 equilibrium selection in stag hunt games.
\newblock \emph{Frontiers of game theory}, pp.\  237, 1993.

\bibitem[Duan et~al.(2024)Duan, Zhang, Diffenderfer, Kailkhura, Sun, Stengel-Eskin, Bansal, Chen, and Xu]{duan2024gtbench}
Jinhao Duan, Renming Zhang, James Diffenderfer, Bhavya Kailkhura, Lichao Sun, Elias Stengel-Eskin, Mohit Bansal, Tianlong Chen, and Kaidi Xu.
\newblock Gtbench: Uncovering the strategic reasoning limitations of llms via game-theoretic evaluations.
\newblock \emph{ArXiv preprint}, abs/2402.12348, 2024.
\newblock URL \url{https://arxiv.org/abs/2402.12348}.

\bibitem[Edelman et~al.(2007)Edelman, Ostrovsky, and Schwarz]{edelman2007internet}
Benjamin Edelman, Michael Ostrovsky, and Michael Schwarz.
\newblock Internet advertising and the generalized second-price auction: Selling billions of dollars worth of keywords.
\newblock \emph{American economic review}, 97\penalty0 (1):\penalty0 242--259, 2007.

\bibitem[Fan et~al.(2023)Fan, Chen, Jin, and He]{Fan2023CanLL}
Caoyun Fan, Jindou Chen, Yaohui Jin, and Hao He.
\newblock Can large language models serve as rational players in game theory? a systematic analysis.
\newblock \emph{ArXiv preprint}, abs/2312.05488, 2023.
\newblock URL \url{https://arxiv.org/abs/2312.05488}.

\bibitem[Feng et~al.(2025)Feng, Dou, Li, Wang, Wang, Guo, Ma, and Kong]{feng2024survey}
Xiachong Feng, Longxu Dou, Ella Li, Qinghao Wang, Haochuan Wang, Yu~Guo, Chang Ma, and Lingpeng Kong.
\newblock A survey on large language model-based social agents in game-theoretic scenarios.
\newblock \emph{arXiv preprint arXiv:2412.03920}, 2025.

\bibitem[Forsythe et~al.(1994)Forsythe, Horowitz, Savin, and Sefton]{forsythe1994fairness}
Robert Forsythe, Joel~L Horowitz, Nathan~E Savin, and Martin Sefton.
\newblock Fairness in simple bargaining experiments.
\newblock \emph{Games and Economic behavior}, 6\penalty0 (3):\penalty0 347--369, 1994.

\bibitem[Fourrier et~al.(2024)Fourrier, Habib, Lozovskaya, Szafer, and Wolf]{open-llm-leaderboard-v2}
Clémentine Fourrier, Nathan Habib, Alina Lozovskaya, Konrad Szafer, and Thomas Wolf.
\newblock Open llm leaderboard v2.
\newblock \url{https://huggingface.co/spaces/open-llm-leaderboard/open_llm_leaderboard}, 2024.

\bibitem[Gandhi et~al.(2023)Gandhi, Sadigh, and Goodman]{gandhi2023strategic}
Kanishk Gandhi, Dorsa Sadigh, and Noah Goodman.
\newblock Strategic reasoning with language models.
\newblock In \emph{NeurIPS 2023 Foundation Models for Decision Making Workshop}, 2023.

\bibitem[Gintis(2014)]{gintis2014bounds}
Herbert Gintis.
\newblock \emph{The bounds of reason: Game theory and the unification of the behavioral sciences-revised edition}.
\newblock Princeton university press, 2014.

\bibitem[Guo et~al.(2025{\natexlab{a}})Guo, Yang, Zhang, Song, Zhang, Xu, Zhu, Ma, Wang, Bi, et~al.]{guo2025deepseek}
Daya Guo, Dejian Yang, Haowei Zhang, Junxiao Song, Ruoyu Zhang, Runxin Xu, Qihao Zhu, Shirong Ma, Peiyi Wang, Xiao Bi, et~al.
\newblock Deepseek-r1: Incentivizing reasoning capability in llms via reinforcement learning.
\newblock \emph{arXiv preprint arXiv:2501.12948}, 2025{\natexlab{a}}.

\bibitem[Guo et~al.(2025{\natexlab{b}})Guo, Wu, Zhu, Leng, Shi, Chen, Fan, Wang, Jiang, Wang, Chen, Huang, Lei, Yuan, Luo, Liu, Ye, Qian, Yan, Zhao, Peng, Li, Yuan, Wu, Cheng, Liu, Wang, Zeng, Liu, Qin, Ding, Xiao, Zhang, Zhang, Xiong, Peng, Chen, Li, Hu, Lin, Hu, Zhang, Wu, Li, Liu, Ling, Qin, Wang, He, Zhang, Yi, Liao, Huang, Zhang, Deng, Deng, Lin, Yuan, Li, Gou, Lou, Wei, Liu, Li, Zhu, Zhong, Li, Zhang, Wu, Li, Xiao, Lin, Yang, Wang, Ji, Hao, Shen, Li, Li, Wu, Zhu, Jiao, Feng, Chen, Duan, Liu, Zeng, Tang, Sun, Chen, Long, Feng, Zhan, Fang, Lu, Hua, Liu, Shen, Zhang, Shen, Wang, Pan, Zhang, Li, Li, Li, Shi, Han, Xiang, Chen, Chen, Li, Yan, Chi, Liu, Du, Wang, Pan, Chen, Chen, Wu, Yuan, Shuai, Tao, Zheng, Zhang, Zhang, Wang, Yang, Zhao, Xu, Liang, Yan, Zhong, Cao, Wu, Liu, Chang, Cai, Ao, Yang, Zhang, Zhong, Jia, Weng, Yu, Huang, Zhu, Yang, Wang, Long, Yin, Li, Zhu, Jia, Zhang, Liu, Zhang, Yang, Luo, Chen, Zhong, Xiao, Li, Wu, Wen, Du, Zhang, Ye, Wu, Liu, Yue, Zhou, Yuan, Xu, Yang, Zhang, Fang, Li, Ren,
  Xiong, Hong, Wang, Sun, Wang, Cai, Zha, An, Zhao, Xu, Chen, Wu, Zheng, Wang, Huang, Zhu, and Song]{guo2025seed15vltechnicalreport}
Dong Guo, Faming Wu, Feida Zhu, Fuxing Leng, Guang Shi, Haobin Chen, Haoqi Fan, Jian Wang, Jianyu Jiang, Jiawei Wang, Jingji Chen, Jingjia Huang, Kang Lei, Liping Yuan, Lishu Luo, Pengfei Liu, Qinghao Ye, Rui Qian, Shen Yan, Shixiong Zhao, Shuai Peng, Shuangye Li, Sihang Yuan, Sijin Wu, Tianheng Cheng, Weiwei Liu, Wenqian Wang, Xianhan Zeng, Xiao Liu, Xiaobo Qin, Xiaohan Ding, Xiaojun Xiao, Xiaoying Zhang, Xuanwei Zhang, Xuehan Xiong, Yanghua Peng, Yangrui Chen, Yanwei Li, Yanxu Hu, Yi~Lin, Yiyuan Hu, Yiyuan Zhang, Youbin Wu, Yu~Li, Yudong Liu, Yue Ling, Yujia Qin, Zanbo Wang, Zhiwu He, Aoxue Zhang, Bairen Yi, Bencheng Liao, Can Huang, Can Zhang, Chaorui Deng, Chaoyi Deng, Cheng Lin, Cheng Yuan, Chenggang Li, Chenhui Gou, Chenwei Lou, Chengzhi Wei, Chundian Liu, Chunyuan Li, Deyao Zhu, Donghong Zhong, Feng Li, Feng Zhang, Gang Wu, Guodong Li, Guohong Xiao, Haibin Lin, Haihua Yang, Haoming Wang, Heng Ji, Hongxiang Hao, Hui Shen, Huixia Li, Jiahao Li, Jialong Wu, Jianhua Zhu, Jianpeng Jiao, Jiashi Feng, Jiaze
  Chen, Jianhui Duan, Jihao Liu, Jin Zeng, Jingqun Tang, Jingyu Sun, Joya Chen, Jun Long, Junda Feng, Junfeng Zhan, Junjie Fang, Junting Lu, Kai Hua, Kai Liu, Kai Shen, Kaiyuan Zhang, Ke~Shen, Ke~Wang, Keyu Pan, Kun Zhang, Kunchang Li, Lanxin Li, Lei Li, Lei Shi, Li~Han, Liang Xiang, Liangqiang Chen, Lin Chen, Lin Li, Lin Yan, Liying Chi, Longxiang Liu, Mengfei Du, Mingxuan Wang, Ningxin Pan, Peibin Chen, Pengfei Chen, Pengfei Wu, Qingqing Yuan, Qingyao Shuai, Qiuyan Tao, Renjie Zheng, Renrui Zhang, Ru~Zhang, Rui Wang, Rui Yang, Rui Zhao, Shaoqiang Xu, Shihao Liang, Shipeng Yan, Shu Zhong, Shuaishuai Cao, Shuangzhi Wu, Shufan Liu, Shuhan Chang, Songhua Cai, Tenglong Ao, Tianhao Yang, Tingting Zhang, Wanjun Zhong, Wei Jia, Wei Weng, Weihao Yu, Wenhao Huang, Wenjia Zhu, Wenli Yang, Wenzhi Wang, Xiang Long, XiangRui Yin, Xiao Li, Xiaolei Zhu, Xiaoying Jia, Xijin Zhang, Xin Liu, Xinchen Zhang, Xinyu Yang, Xiongcai Luo, Xiuli Chen, Xuantong Zhong, Xuefeng Xiao, Xujing Li, Yan Wu, Yawei Wen, Yifan Du, Yihao Zhang,
  Yining Ye, Yonghui Wu, Yu~Liu, Yu~Yue, Yufeng Zhou, Yufeng Yuan, Yuhang Xu, Yuhong Yang, Yun Zhang, Yunhao Fang, Yuntao Li, Yurui Ren, Yuwen Xiong, Zehua Hong, Zehua Wang, Zewei Sun, Zeyu Wang, Zhao Cai, Zhaoyue Zha, Zhecheng An, Zhehui Zhao, Zhengzhuo Xu, Zhipeng Chen, Zhiyong Wu, Zhuofan Zheng, Zihao Wang, Zilong Huang, Ziyu Zhu, and Zuquan Song.
\newblock Seed1.5-vl technical report, 2025{\natexlab{b}}.
\newblock URL \url{https://arxiv.org/abs/2505.07062}.

\bibitem[Guo(2023)]{guo2023gpt}
Fulin Guo.
\newblock Gpt in game theory experiments.
\newblock \emph{ArXiv preprint}, abs/2305.05516, 2023.
\newblock URL \url{https://arxiv.org/abs/2305.05516}.

\bibitem[Guo et~al.(2023)Guo, Yang, Yoo, Lin, Iwasawa, and Matsuo]{guo2023suspicion}
Jiaxian Guo, Bo~Yang, Paul Yoo, Bill~Yuchen Lin, Yusuke Iwasawa, and Yutaka Matsuo.
\newblock Suspicion-agent: Playing imperfect information games with theory of mind aware gpt-4.
\newblock \emph{ArXiv preprint}, abs/2309.17277, 2023.
\newblock URL \url{https://arxiv.org/abs/2309.17277}.

\bibitem[Herr et~al.(2024)Herr, Acero, Raileanu, P{\'e}rez-Ortiz, and Li]{herr2024large}
Nathan Herr, Fernando Acero, Roberta Raileanu, Mar{\'\i}a P{\'e}rez-Ortiz, and Zhibin Li.
\newblock Are large language models strategic decision makers? a study of performance and bias in two-player non-zero-sum games.
\newblock \emph{ArXiv preprint}, abs/2407.04467, 2024.
\newblock URL \url{https://arxiv.org/abs/2407.04467}.

\bibitem[Horton(2023)]{horton2023large}
John~J Horton.
\newblock Large language models as simulated economic agents: What can we learn from homo silicus?
\newblock Technical report, National Bureau of Economic Research, 2023.

\bibitem[Hua et~al.(2024)Hua, Liu, Li, Amayuelas, Chen, Jiang, Jin, Fan, Sun, Wang, et~al.]{hua2024game}
Wenyue Hua, Ollie Liu, Lingyao Li, Alfonso Amayuelas, Julie Chen, Lucas Jiang, Mingyu Jin, Lizhou Fan, Fei Sun, William Wang, et~al.
\newblock Game-theoretic llm: Agent workflow for negotiation games.
\newblock \emph{arXiv preprint arXiv:2411.05990}, 2024.

\bibitem[Huang et~al.(2024)Huang, Li, Lam, Liang, Wang, Yuan, Jiao, Wang, Tu, and Lyu]{huang2024far}
Jen-tse Huang, Eric~John Li, Man~Ho Lam, Tian Liang, Wenxuan Wang, Youliang Yuan, Wenxiang Jiao, Xing Wang, Zhaopeng Tu, and Michael~R Lyu.
\newblock How far are we on the decision-making of llms? evaluating llms' gaming ability in multi-agent environments.
\newblock \emph{ArXiv preprint}, abs/2403.11807, 2024.
\newblock URL \url{https://arxiv.org/abs/2403.11807}.

\bibitem[Imani et~al.(2023)Imani, Du, and Shrivastava]{imani2023mathprompter}
Shima Imani, Liang Du, and Harsh Shrivastava.
\newblock Mathprompter: Mathematical reasoning using large language models.
\newblock \emph{ArXiv preprint}, abs/2303.05398, 2023.
\newblock URL \url{https://arxiv.org/abs/2303.05398}.

\bibitem[Jaech et~al.(2024)Jaech, Kalai, Lerer, Richardson, El-Kishky, Low, Helyar, Madry, Beutel, Carney, et~al.]{jaech2024openai}
Aaron Jaech, Adam Kalai, Adam Lerer, Adam Richardson, Ahmed El-Kishky, Aiden Low, Alec Helyar, Aleksander Madry, Alex Beutel, Alex Carney, et~al.
\newblock Openai o1 system card.
\newblock \emph{arXiv preprint arXiv:2412.16720}, 2024.

\bibitem[K{\i}c{\i}man et~al.(2023)K{\i}c{\i}man, Ness, Sharma, and Tan]{kiciman2023causal}
Emre K{\i}c{\i}man, Robert Ness, Amit Sharma, and Chenhao Tan.
\newblock Causal reasoning and large language models: Opening a new frontier for causality.
\newblock \emph{ArXiv preprint}, abs/2305.00050, 2023.
\newblock URL \url{https://arxiv.org/abs/2305.00050}.

\bibitem[Kosinski(2023)]{kosinski2023theory}
Michal Kosinski.
\newblock Theory of mind might have spontaneously emerged in large language models.
\newblock \emph{ArXiv preprint}, abs/2302.02083, 2023.
\newblock URL \url{https://arxiv.org/abs/2302.02083}.

\bibitem[Kreps(1990)]{kreps1990game}
David~M Kreps.
\newblock \emph{Game theory and economic modelling}.
\newblock Oxford University Press, 1990.

\bibitem[Lei et~al.(2023)Lei, Liao, Ding, et~al.]{lei2023boosting}
Bin Lei, Chunhua Liao, Caiwen Ding, et~al.
\newblock Boosting logical reasoning in large language models through a new framework: The graph of thought.
\newblock \emph{ArXiv preprint}, abs/2308.08614, 2023.
\newblock URL \url{https://arxiv.org/abs/2308.08614}.

\bibitem[Li et~al.(2023)Li, Li, and Liu]{li2023beyond}
Jiatong Li, Rui Li, and Qi~Liu.
\newblock Beyond static datasets: A deep interaction approach to llm evaluation.
\newblock \emph{ArXiv preprint}, abs/2309.04369, 2023.
\newblock URL \url{https://arxiv.org/abs/2309.04369}.

\bibitem[Liddle \& Nettle(2006)Liddle and Nettle]{liddle2006higher}
Bethany Liddle and Daniel Nettle.
\newblock Higher-order theory of mind and social competence in school-age children.
\newblock \emph{Journal of Cultural and Evolutionary Psychology}, 4\penalty0 (3-4):\penalty0 231--244, 2006.

\bibitem[Liu et~al.(2024)Liu, Anand, Zhou, Huang, and Zhao]{liu2024interintent}
Ziyi Liu, Abhishek Anand, Pei Zhou, Jen-tse Huang, and Jieyu Zhao.
\newblock Interintent: Investigating social intelligence of llms via intention understanding in an interactive game context.
\newblock \emph{ArXiv preprint}, abs/2406.12203, 2024.
\newblock URL \url{https://arxiv.org/abs/2406.12203}.

\bibitem[Lor{\`e} \& Heydari(2023)Lor{\`e} and Heydari]{lore2023strategic}
Nunzio Lor{\`e} and Babak Heydari.
\newblock Strategic behavior of large language models: Game structure vs. contextual framing.
\newblock \emph{ArXiv preprint}, abs/2309.05898, 2023.
\newblock URL \url{https://arxiv.org/abs/2309.05898}.

\bibitem[Phelps \& Russell(2023)Phelps and Russell]{phelps2023investigating}
Steve Phelps and Yvan~I. Russell.
\newblock The machine psychology of cooperation: Can gpt models operationalise prompts for altruism, cooperation, competitiveness and selfishness in economic games?
\newblock \emph{ArXiv preprint}, 2023.
\newblock URL \url{https://api.semanticscholar.org/CorpusID:258685424}.

\bibitem[Premack \& Woodruff(1978)Premack and Woodruff]{premack1988does}
David Premack and Guy Woodruff.
\newblock Does the chimpanzee have a theory of mind?
\newblock \emph{Behavioral and brain sciences}, 1\penalty0 (4):\penalty0 515--526, 1978.

\bibitem[Robinson \& Goforth(2005)Robinson and Goforth]{robinson2005topology}
David Robinson and David Goforth.
\newblock \emph{The topology of the 2x2 games: a new periodic table}, volume~3.
\newblock Psychology Press, 2005.

\bibitem[Shao et~al.(2024)Shao, Wang, Zhu, Xu, Song, Bi, Zhang, Zhang, Li, Wu, et~al.]{shao2024deepseekmath}
Zhihong Shao, Peiyi Wang, Qihao Zhu, Runxin Xu, Junxiao Song, Xiao Bi, Haowei Zhang, Mingchuan Zhang, YK~Li, Y~Wu, et~al.
\newblock Deepseekmath: Pushing the limits of mathematical reasoning in open language models.
\newblock \emph{arXiv preprint arXiv:2402.03300}, 2024.

\bibitem[Shapira et~al.(2024)Shapira, Madmon, Reinman, Amouyal, Reichart, and Tennenholtz]{shapira2024glee}
Eilam Shapira, Omer Madmon, Itamar Reinman, Samuel~Joseph Amouyal, Roi Reichart, and Moshe Tennenholtz.
\newblock Glee: A unified framework and benchmark for language-based economic environments.
\newblock \emph{arXiv preprint arXiv:2410.05254}, 2024.

\bibitem[Team et~al.(2025)Team, Kamath, Ferret, Pathak, Vieillard, Merhej, Perrin, Matejovicova, Ram{\'e}, Rivi{\`e}re, et~al.]{team2025gemma}
Gemma Team, Aishwarya Kamath, Johan Ferret, Shreya Pathak, Nino Vieillard, Ramona Merhej, Sarah Perrin, Tatiana Matejovicova, Alexandre Ram{\'e}, Morgane Rivi{\`e}re, et~al.
\newblock Gemma 3 technical report.
\newblock \emph{arXiv preprint arXiv:2503.19786}, 2025.

\bibitem[Toroghi et~al.(2024)Toroghi, Guo, Pesaranghader, and Sanner]{toroghi-etal-2024-verifiable}
Armin Toroghi, Willis Guo, Ali Pesaranghader, and Scott Sanner.
\newblock Verifiable, debuggable, and repairable commonsense logical reasoning via {LLM}-based theory resolution.
\newblock In Yaser Al-Onaizan, Mohit Bansal, and Yun-Nung Chen (eds.), \emph{Proceedings of the 2024 Conference on Empirical Methods in Natural Language Processing}, pp.\  6634--6652, Miami, Florida, USA, November 2024. Association for Computational Linguistics.
\newblock \doi{10.18653/v1/2024.emnlp-main.379}.
\newblock URL \url{https://aclanthology.org/2024.emnlp-main.379/}.

\bibitem[Tsebelis(1990)]{tsebelis1990nested}
George Tsebelis.
\newblock \emph{Nested Games: Rational Choice in Comparative Politics}.
\newblock University of California Press, 1990.

\bibitem[Ullman(2023)]{ullman2023large}
Tomer Ullman.
\newblock Large language models fail on trivial alterations to theory-of-mind tasks.
\newblock \emph{ArXiv preprint}, abs/2302.08399, 2023.
\newblock URL \url{https://arxiv.org/abs/2302.08399}.

\bibitem[Wang et~al.(2024)Wang, Yao, Xu, Qiao, Deng, Wang, Chen, Gu, Jiang, Xie, et~al.]{wang2024knowledge}
Mengru Wang, Yunzhi Yao, Ziwen Xu, Shuofei Qiao, Shumin Deng, Peng Wang, Xiang Chen, Jia-Chen Gu, Yong Jiang, Pengjun Xie, et~al.
\newblock Knowledge mechanisms in large language models: A survey and perspective.
\newblock \emph{arXiv preprint arXiv:2407.15017}, 2024.

\bibitem[Wei et~al.(2022)Wei, Wang, Schuurmans, Bosma, Xia, Chi, Le, Zhou, et~al.]{wei2022chain}
Jason Wei, Xuezhi Wang, Dale Schuurmans, Maarten Bosma, Fei Xia, Ed~Chi, Quoc~V Le, Denny Zhou, et~al.
\newblock Chain-of-thought prompting elicits reasoning in large language models.
\newblock \emph{Advances in neural information processing systems}, 35:\penalty0 24824--24837, 2022.

\bibitem[Xu et~al.(2023)Xu, Hu, Zhou, Ren, Dong, Keutzer, Ng, and Feng]{xu2023magic}
Lin Xu, Zhiyuan Hu, Daquan Zhou, Hongyu Ren, Zhen Dong, Kurt Keutzer, See-Kiong Ng, and Jiashi Feng.
\newblock Magic: Investigation of large language model powered multi-agent in cognition, adaptability, rationality and collaboration.
\newblock In \emph{ICLR 2024 Workshop on Large Language Model (LLM) Agents}, 2023.

\bibitem[Yang et~al.(2024)Yang, Yang, Zhang, Hui, Zheng, Yu, Li, Liu, Huang, Wei, Lin, Yang, Tu, Zhang, Yang, Yang, Zhou, Lin, Dang, Lu, Bao, Yang, Yu, Li, Xue, Zhang, Zhu, Men, Lin, Li, Xia, Ren, Ren, Fan, Su, Zhang, Wan, Liu, Cui, Zhang, and Qiu]{qwen2.5}
An~Yang, Baosong Yang, Beichen Zhang, Binyuan Hui, Bo~Zheng, Bowen Yu, Chengyuan Li, Dayiheng Liu, Fei Huang, Haoran Wei, Huan Lin, Jian Yang, Jianhong Tu, Jianwei Zhang, Jianxin Yang, Jiaxi Yang, Jingren Zhou, Junyang Lin, Kai Dang, Keming Lu, Keqin Bao, Kexin Yang, Le~Yu, Mei Li, Mingfeng Xue, Pei Zhang, Qin Zhu, Rui Men, Runji Lin, Tianhao Li, Tingyu Xia, Xingzhang Ren, Xuancheng Ren, Yang Fan, Yang Su, Yichang Zhang, Yu~Wan, Yuqiong Liu, Zeyu Cui, Zhenru Zhang, and Zihan Qiu.
\newblock Qwen2.5 technical report.
\newblock \emph{arXiv preprint arXiv:2412.15115}, 2024.

\bibitem[Yim et~al.(2024)Yim, Chan, Shi, Deng, Fan, Zheng, and Song]{yim2024evaluating}
Yauwai Yim, Chunkit Chan, Tianyu Shi, Zheye Deng, Wei Fan, Tianshi Zheng, and Yangqiu Song.
\newblock Evaluating and enhancing llms agent based on theory of mind in guandan: A multi-player cooperative game under imperfect information.
\newblock \emph{ArXiv preprint}, abs/2408.02559, 2024.
\newblock URL \url{https://arxiv.org/abs/2408.02559}.

\bibitem[Zhang et~al.(2024{\natexlab{a}})Zhang, Mao, Ge, Wang, de~Wynter, Xia, Wu, Song, Lan, and Wei]{zhang2024llm}
Yadong Zhang, Shaoguang Mao, Tao Ge, Xun Wang, Adrian de~Wynter, Yan Xia, Wenshan Wu, Ting Song, Man Lan, and Furu Wei.
\newblock Llm as a mastermind: A survey of strategic reasoning with large language models.
\newblock \emph{ArXiv preprint}, abs/2404.01230, 2024{\natexlab{a}}.
\newblock URL \url{https://arxiv.org/abs/2404.01230}.

\bibitem[Zhang et~al.(2024{\natexlab{b}})Zhang, Mao, Ge, Wang, Xia, Lan, and Wei]{zhang2024k}
Yadong Zhang, Shaoguang Mao, Tao Ge, Xun Wang, Yan Xia, Man Lan, and Furu Wei.
\newblock K-level reasoning with large language models.
\newblock \emph{ArXiv preprint}, abs/2402.01521, 2024{\natexlab{b}}.
\newblock URL \url{https://arxiv.org/abs/2402.01521}.

\bibitem[Zhao et~al.(2024)Zhao, Lee, and Hsu]{zhao2024large}
Zirui Zhao, Wee~Sun Lee, and David Hsu.
\newblock Large language models as commonsense knowledge for large-scale task planning.
\newblock \emph{Advances in Neural Information Processing Systems}, 36, 2024.

\end{thebibliography}
